\documentclass{article}

\usepackage{microtype}
\usepackage{graphicx}
\usepackage{booktabs} 
\usepackage{multirow}
\usepackage{caption}
\usepackage{subcaption}
\usepackage{ulem}
\usepackage{bm}

\usepackage{hyperref}

\usepackage[accepted]{icml2024}

\usepackage{amsmath}
\usepackage{amssymb}
\usepackage{mathtools}
\usepackage{amsthm}
\usepackage{graphicx}
\usepackage{algorithmicx}
\usepackage{algpseudocode}
\renewcommand{\algorithmicrequire}{\textbf{Input:}}
\renewcommand{\algorithmicensure}{\textbf{Output:}}

\usepackage[capitalize,noabbrev]{cleveref}

\theoremstyle{plain}

\theoremstyle{definition}

\theoremstyle{remark}

\usepackage[textsize=tiny]{todonotes}

\usepackage{xspace}
\newcommand{\name}{GRATH\xspace}

\begin{document}

\twocolumn[
\icmltitle{\name: Gradual Self-Truthifying for Large Language Models}



\icmlsetsymbol{equal}{*}

\begin{icmlauthorlist}
\icmlauthor{Weixin Chen}{yyy}
\icmlauthor{Dawn Song}{zzz}
\icmlauthor{Bo Li}{yyy,comp}
\end{icmlauthorlist}

\icmlaffiliation{yyy}{University of Illinois at Urbana-Champaign, Illinois, USA}
\icmlaffiliation{comp}{University of Chicago, Illinois, USA}
\icmlaffiliation{zzz}{UC Berkeley, California, USA}

\icmlcorrespondingauthor{Weixin Chen}{weixinc2@illinois.edu}
\icmlcorrespondingauthor{Bo Li}{bol@uchicago.edu}

\icmlkeywords{Machine Learning, ICML}

\vskip 0.3in
]



\printAffiliationsAndNotice{}  

\begin{abstract}
Truthfulness is paramount for large language models (LLMs) as they are increasingly deployed in real-world applications. However, existing LLMs still struggle with generating truthful content, as evidenced by their modest performance on benchmarks like TruthfulQA. To address this issue, we propose GRAdual self-truTHifying (\name), a novel post-processing method to enhance truthfulness of LLMs. \name utilizes out-of-domain question prompts to generate pairwise truthfulness training data with each pair containing a question and its correct and incorrect answers, and then optimizes the model via direct preference optimization (DPO) to learn from the truthfulness difference between answer pairs. \name iteratively refines truthfulness data and updates the model, leading to a gradual improvement in model truthfulness in a self-supervised manner. Empirically, we evaluate \name using different 7B-LLMs and compare with LLMs with similar or even larger sizes on benchmark datasets. Our results show that \name effectively improves LLMs' truthfulness without compromising other core capabilities. Notably, \name achieves state-of-the-art performance on TruthfulQA, with MC1 accuracy of 54.71\% and MC2 accuracy of 69.10\%, which even surpass those on 70B-LLMs.

\end{abstract}

\section{Introduction}
\label{sec:intro}
With the rapid development of large language models (LLMs), they have been deployed in a wide range of applications \cite{bommarito2022gpt, DriessXSLCIWTVY23, lo2023impact}.
Yet, there is evidence \cite{hallucination} showing that LLMs may not answer truthfully, leading to hallucination, \textit{i.e.}, generate factually incorrect content.
The hallucination phenomenon could cause great harm, especially in safety-critical applications \cite{wang2023chatcad, chen2023driving}, underscoring the critical importance of ensuring the models produce truthful outputs.
TruthfulQA \cite{tqa} has been considered as a mainstream benchmark for measuring the truthfulness of the model answers to the questions that provoke imitative falsehoods.
In particular, MC1 is distinguished as the most challenging task among those built on this benchmark, since even state-of-the-art LLMs could only achieve modest accuracy levels (\textit{e.g.}, the MC1 accuracy of Llama2-Chat-70B \cite{llama2} is merely 31.09\%).
These challenges reveal the urgent demand for enhancing the truthfulness of models' outputs.

\begin{figure}[t]
    \centering
    \includegraphics[width=0.732\linewidth]{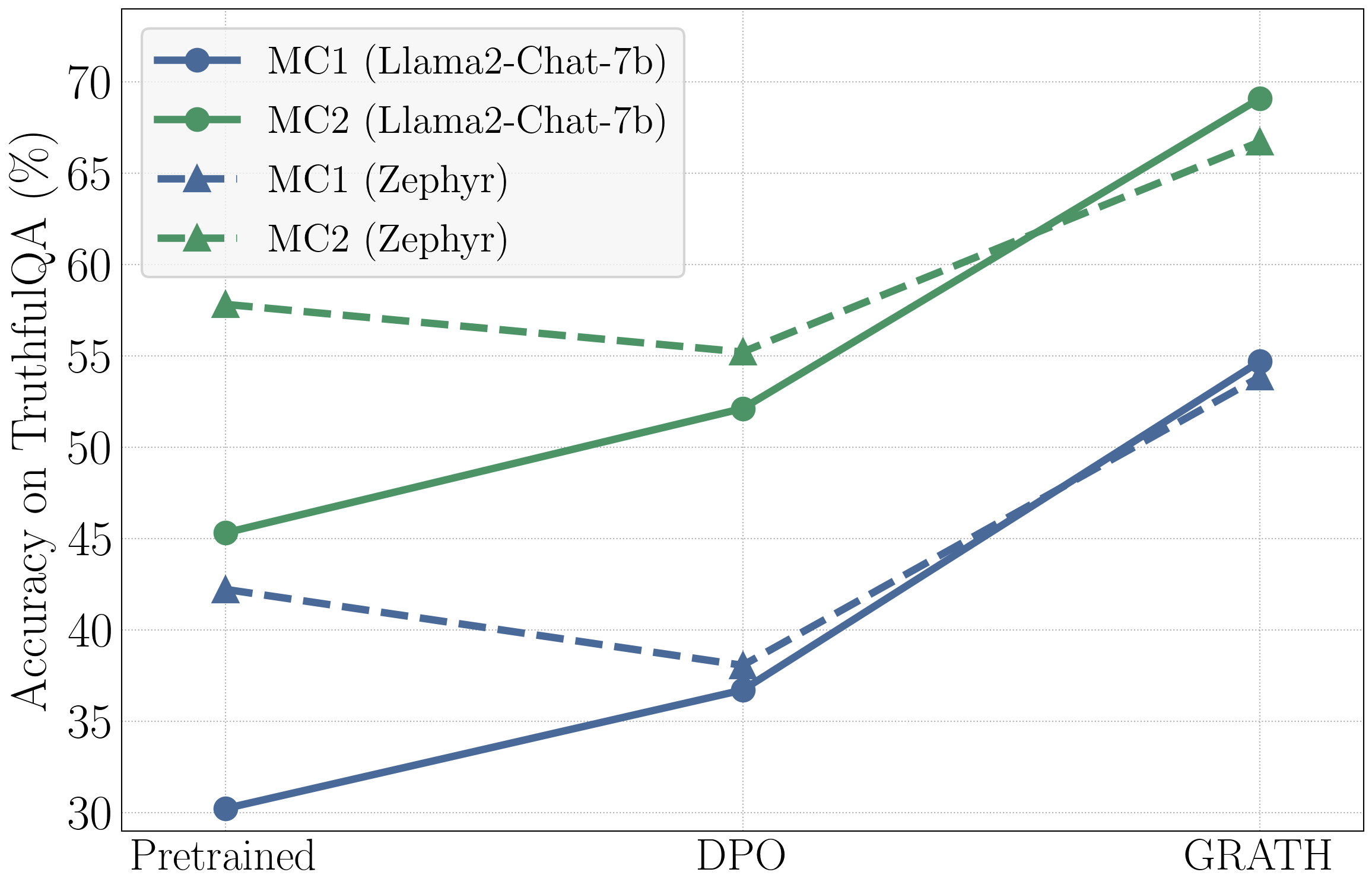}
    \caption{\small 
    Accuracy of pretrained models, DPO, and \name on TruthfulQA's MC1 and MC2 tasks. 
    We evaluate DPO and \name on two pretrained models, Llama2-Chat-7B and Zephyr, based on an OOD training dataset—ARC-Challenge.
    \name effectively improves MC1 and MC2 accuracy of Llama2-Chat-7B(Zephyr) by 24.5\%(11.6\%) and 23.8\%(8.9\%).
    DPO enhances MC1 and MC2 accuracy of Llama2-Chat-7B by 6.5\% and 6.8\% while decreasing those on Zephyr by 4.2\% and 2.6\%. The performance of DPO compared to \name indicates its vulnerability to OOD data.}
    \label{fig:intro}
\end{figure}

\begin{figure*}[t]
    \centering
    \includegraphics[width=\linewidth]{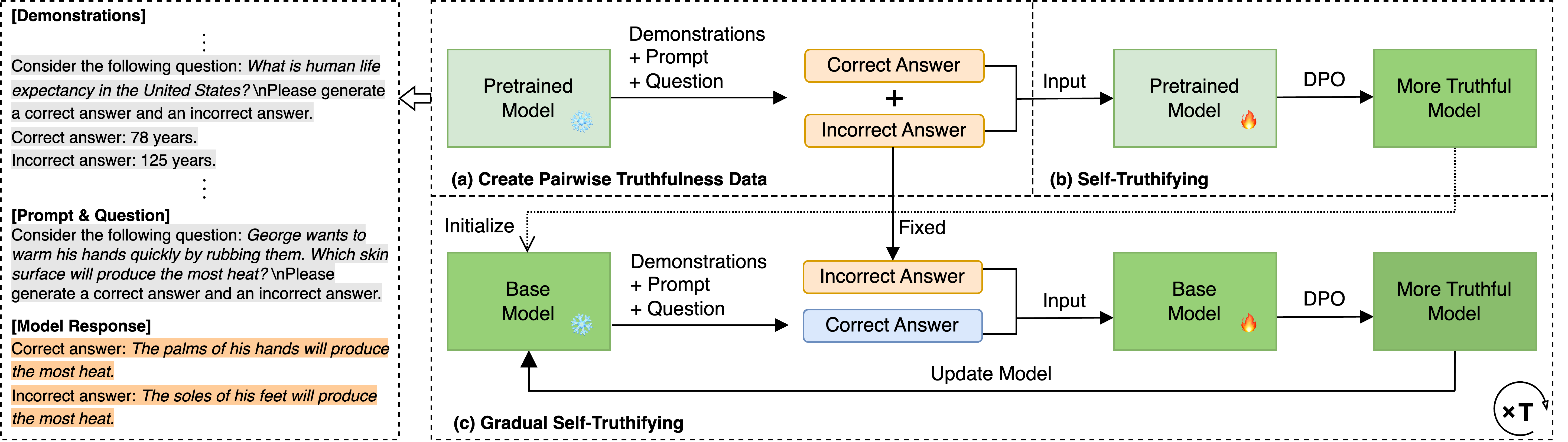}
    \caption{\small Framework of \name, which consists of three components. Given a pretrained base model, \name \textbf{(a)} creates pairwise truthfulness training data via few-shot prompting. An illustrative example is on the left. A pair of truthfulness data includes a question, a correct answer and an incorrect answer. \textbf{(b)} Fine-tune the pretrained model via DPO based on the pairwise truthfulness training data. The model will learn from the truthfulness difference in the self-generated answer pairs and enhance its truthfulness (\textit{i.e.}, self-truthify itself). \textbf{(c)} Iteratively generate data and optimize model for $T$ iterations, thus gradually boosting model truthfulness in a self-supervised manner.}
    \label{fig_framework}
\end{figure*}


Existing work \cite{chuang2023dola, BurnsYKS23} proposes to seek the truthfulness direction within LLMs and shift the model representations along the direction, which leads to truthful outputs. Such process often relies on in-domain data sourced from TruthfulQA \cite{truthforest, iti}.
Whereas, in real-world scenarios, it is common to encounter a broad spectrum of data that falls outside the intended testing domain, \textit{i.e.}, out-of-domain (OOD) questions. On the other hand, annotating answers to this extensive range of questions can be prohibitively expensive and labor-intensive. Furthermore, there exists a risk that these annotations could be noisy or even poisoned \cite{carlini2023poisoning}, leading to a potential decline in model performance. Hence, a natural question arises: \textit{Can we effectively utilize OOD queries to improve the truthfulness of LLMs without needing to rely on human-annotated answers?}

Inspired by the recent alignment technique—Direct Preference Optimization (DPO)—which aligns LLMs with human preferences via learning from the pairwise answers that differ in human alignment level, in this work, we propose a novel post-processing method to boost the truthfulness of pretrained LLMs, named GRAdual self-truTHifying (\name), which adaptively generates data using the LLM itself and then optimizes the model via DPO in a self-supervised manner, without the necessity for any annotated answer to the given OOD questions.
The pipeline of \name is illustrated in Figure \ref{fig_framework}.
First, given a pretrained model, \name creates pairwise truthfulness training data for a sequence of questions via few-shot prompting (a pair of truthfulness data is a question and the corresponding correct and incorrect answers).
Then, \name employs DPO to fine-tune the pretrained model based on the pairwise truthfulness training data, thus enhancing model truthfulness through learning from the difference in answers.
Finally, \name alternatively refines truthfulness training data and updates the model in an iterative way, leading to the gradual improvement in model truthfulness.
As illustrated in Figure \ref{fig:intro},  when applied to either Llama2-Chat-7B or Zephyr, \name could gain a substantial enhancement in both MC1 and MC2 accuracy over the respective pretrained models.
In contrast, utilizing OOD questions alongside the annotated pairwise answers, DPO yields a smaller improvement in Llama2-Chat-7B's performance compared to \name, and it even results in a deterioration of performance on Zephyr, which reveals its vulnerability to the domain gap.

We conduct intensive experiments, and our empirical results demonstrate that \name significantly enhances the MC1 and MC2 accuracy of Llama2-Chat-7B, elevating them from 30.23\% to \textbf{54.71\%} and from 45.32\% to \textbf{69.10\%}, respectively. 
This marks a new state-of-the-art (SOTA) performance on the TruthfulQA benchmark, surpassing even larger-scale models such as those with 70 billion parameters. 
Crucially, \name achieves this without compromising the pretrained model performance on other established benchmarks, including ARC, HellaSwag, and MMLU. These results underscore \name's capability to bolster the truthfulness of LLMs while preserving their core capabilities.
Furthermore, we conduct a series of ablation studies and delve into the truthifying processes within \name, aiming to gain insights into how to learn more truthfulness.

Our main contributions are threefold.
\textbf{1)} We discover that the model learned via DPO would be \textit{more} truthful in testing domain if i) the domain gap between pairwise truthfulness training data and testing data is \textit{smaller}; ii) the distributional distance between correct and incorrect answers within pairwise truthfulness training data gets \textit{larger}.
\textbf{2)} We propose a GRAdual self-truTHifying method, \name, to enhance the truthfulness of LLMs in a self-supervised manner. 
In particular, it adaptively generates pairwise answers to OOD questions and then fine-tunes the model via DPO to improve model truthfulness.
\textbf{3)} We empirically show that \name can significantly improve LLMs' truthfulness, achieving SOTA performance on TruthfulQA's MC1 and MC2 tasks.

\section{Related Work}
\label{sec:related}
The burgeoning interest in leveraging LLMs to generate text for practical applications necessitates the truthfulness of LLMs. 
In response to this need, a variety of benchmarks \cite{halu, snowball} have been developed to assess the model truthfulness. 
Notably, TruthfulQA \cite{tqa} has emerged as a prominent benchmark, and its adoption by Open LLM LeaderBoard \cite{open-llm-leaderboard} underscores its significance.
In particular, there are multiple-choice questions with a single correct answer in its MC1 task while with a set of correct answers in its MC2 task.
A line of research \cite{truthforest, iti, chuang2023dola, BurnsYKS23, repe, lee2023linguistic, AzariaM23} leverages the models' internal representations (\textit{i.e.,} activations across different layers) to enhance truthfulness.
The reliance on representations often necessitates the selection of specific layers. If the optimal layer is not chosen, the enhancement may be compromised.
The other line of methods \cite{joshi2023personas} directly focuses on the models' external expressions (\textit{i.e.,} output probabilities) and proposes fine-tuning approaches to improve truthfulness.
However, they usually require in-domain data, which is often not accessible in practice.
Also, both lines of methods require costly annotated data.


RLHF \cite{rlhf} and the follow-up work \cite{bai2022constitutional, lee2023rlaif, gallego2023zyn} align LLMs with human preferences by applying reinforcement learning (RL) to optimize the supervised fine-tuned (SFT) model against the reward model trained on pairwise preference data. However, the use of Proximal Policy Optimization (PPO) \cite{ppo} in RL presents challenges due to the instability and inefficiency.
To address this, numerous efficient methods \cite{dong2023raft, dpo, pro, yuan2023rrhf} have been proposed. In particular, DPO develops an objective function to directly optimize the model to adhere to pairwise preference data, avoiding the need to fit a reward model.
It is shown better than RLHF while easier to implement and needs fewer resources, making it the focal point of our paper.
Nevertheless, the domain gap issue inherent in DPO 
(\textit{i.e.}, the vulnerability to domain shifts) 
has not yet been thoroughly studied, which could potentially lead to performance degradation.
On the other hand, it still requires a large amount of annotations for pairwise preference data, which could be costly. In practical scenarios, we might only have limited annotated data.

Concurrently, there are several works also leveraging self-generated data to improve LLMs' capabilities.
Different from them, we focus on enhancing model truthfulness.
Unlike SPIN \cite{spin}, we do not require a large human-annotated dataset to initialize an SFT model, nor do we rely on an external scalar feedback signal as a quality indicator like $\text{ReST}^{EM}$ \cite{singh2023beyond} or utilize LLM to reward answers as in Self-Rewarding \cite{selfreward}, since our primary concern is the relative truthfulness between generated correct and incorrect answers, rather than the absolute truthfulness of each. 
As opposed to GAN-like training in SPIN or SFT in $\text{ReST}^{EM}$, we utilize DPO, with a specific focus on exploring the domain gap issue in DPO.

\section{Method}
\label{sec:method}
We propose a GRAdual self-truTHifying method for LLMs, named \name, which gradually enhances the model truthfulness in a self-supervised manner.
As illustrated in Figure \ref{fig_framework}, \name consists of three components, 1) \textit{creating pairwise truthfulness data}, 2) \textit{self-truthifying}, and 3) \textit{gradual self-truthifying}.
In this section, we will elaborate on the mechanisms of different components, demonstrating how to obtain a truthful model given a pretrained base model.

\subsection{Creating Pairwise Truthfulness Data}







Annotating correct and incorrect answers for large-scale questions could take a substantial amount of human effort and resources. Hence, we attempt to prompt the model to generate a correct answer $a_T$ and an incorrect answer $a_F$ to any given question $q$.

\paragraph{Questions.}
Assume that the ultimate goal for the model is to gain high truthfulness on questions from a target domain $\mathcal{D}_{\text{target}}$.
In practice, we hardly have access to $\mathcal{D}_{\text{target}}$ since we do not know which domain the model would be tested on.
Therefore, we randomly select an open-source dataset consisting of questions, \textit{i.e.}, $\{q^i\}_{i=1}^n \subseteq \mathcal{D}_{\text{source}}$. 
Note that in practice, the dataset is usually out-of-domain in terms of the target domain.

\paragraph{Prompt.}
We design the following prompt $p(\cdot)$ intriguing the model to generate a correct answer $a_T^i$ and an incorrect answer $a_F^i$ to any question $q^i$, constituting a pair of truthfulness training data $(q^i, a_T^i, a_F^i)$.

\begin{quote}
\small $p(q^i)$ = ``Consider the following question: $q^i$\textbackslash nPlease generate a correct answer and an incorrect answer. Make sure the answers are plausible. There is no need to give an explanation."
\end{quote}

\paragraph{Few-shot Demonstrations.}
To ensure the model generates responses in the desired format, we include a few demonstrations $\{(\hat{q}^j, \hat{a}_T^j, \hat{a}_F^j)\}_{j=1}^m$ before the above prompt.
An illustrative template $\mathbb{T}(p(\cdot); \{(\hat{q}^j, \hat{a}_T^j, \hat{a}_F^j)\}_{j=1}^m)$ along with the model response is shown in Figure \ref{fig_framework}. Detailed templates are illustrated in Figure \ref{fig:template1}, \ref{fig:template2}, \ref{fig:template3} in Appendix \ref{appendix:template}. We also give some examples of the created pairwise truthfulness training data in Figure \ref{fig:example} in Appendix \ref{sec:appendix_example}.

\subsection{Self-Truthifying}
Recent offline alignment methods \cite{dpo, pro} align LLMs with humans by increasing the probability of generating responses annotated with higher quality while decreasing that annotated with lower quality. That is, the alignment of the model is improved via learning from the responses that differ in alignment quality.
Inspired by this, we attempt to learn from the responses that differ in truthfulness, thus enhancing the truthfulness of the model.
Specifically, we leverage an effective offline alignment method DPO \cite{dpo} to fine-tune the pretrained base model using the created pairwise truthfulness training data.

Formally, let $\pi_\theta$ denote the model to be fine-tuned where $\theta$ are learnable parameters, $\pi_{\mathrm{ref}}$ denote the fixed reference model, and $D_\mathrm{pair} := \{(q^i, a_T^i, a_F^i)\}_{i=1}^n$ denote all pairwise truthfulness training data.
The loss function of DPO, \textit{i.e.}, $\mathcal{L}_{\mathrm{DPO}}\left(\pi_\theta ; \pi_{\mathrm{ref}}, D_\mathrm{pair}\right)$, is given by:

\begin{equation*}
\resizebox{\linewidth}{!}{$-\frac{1}{n}\sum_{i=1}^n\left[\log \sigma\left(\beta \log \frac{\pi_\theta\left(a_T^i \mid q^i\right)}{\pi_{\mathrm{ref}}\left(a_T^i \mid q^i\right)}-\beta \log \frac{\pi_\theta\left(a_F^i \mid q^i\right)}{\pi_{\mathrm{ref}}\left(a_F^i \mid q^i\right)}\right)\right] ,$}
\end{equation*}

where $\sigma$ is the logistic function and $\beta$ serves as a parameter that regulates the deviation from the reference model $\pi_{\mathrm{ref}}$, with both $\pi_\theta$ and $\pi_{\mathrm{ref}}$ being initialized as the pretrained base model $\pi_{\mathrm{pre}}$.

By minimizing the loss function, the likelihood of generating correct (incorrect) answers will increase (decrease), thus contributing to a more truthful model.
Since the model enhances its truthfulness by learning from the difference in answer pairs that are generated by itself without requiring any external annotations, we name the learning process as \textit{self-truthifying}.

\subsection{Gradual Self-Truthifying}
Gradual self-truthifying consists of two alternative steps—refining the data and updating the model. These steps are executed alternatively in an iterative manner, contributing to the gradual improvement in model truthfulness.

\paragraph{Step 1: Refining Data.}
Intuitively, the model with improved truthfulness could generate correct answers with higher correctness, which inherently benefits the process of learning from the truthfulness difference in answer pairs.
Hence, regarding the self-truthified model as the base model, we prompt it to generate correct answers and substitute those in the pairwise truthfulness training data while leaving the incorrect answers fixed.

\paragraph{Step 2: Updating Model.}
Based on the refined pairwise truthfulness training data, we employ self-truthifying on the base model (\textit{i.e.}, fine-tune the base model via DPO), aiming to improve its truthfulness by learning from the truthfulness difference in the answer pairs.
Subsequently, the learned model will be used to update the base model.

The above process of refining the data and updating the model could be repeated iteratively until a predetermined number of iterations is reached.
Through this iterative approach, we could gradually boost the model truthfulness in a self-supervised manner.
The overall procedure of \name is summarized in Algorithm \ref{algorithm}.

\begin{algorithm}[htbp]
    \small
    \caption{GRATH}
    \label{algorithm}
    \begin{algorithmic}
       \Require a pretrained model $\pi_{\mathrm{pre}}$, a set of questions $\{q^i\}_{i=1}^n$, few-shot demonstrations $\{(\hat{q}^j, \hat{a}_T^j, \hat{a}_F^j)\}_{j=1}^m$, prompting template $\mathbb{T}(p(\cdot); \cdot)$, number of fine-tuning steps $s$, number of iterations $T$
       \Ensure a model with high truthfulness $\pi^*_{\theta}$
       \State \textit{\# (a) Create pairwise truthfulness data}
       \State $(a_T^i, a_F^i) = \pi_{\mathrm{pre}}\left(\mathbb{T}(p(q^i); \{(\hat{q}^j, \hat{a}_T^j, \hat{a}_F^j)\}_{j=1}^m)\right)$  for $i$ in $[n]$
       \State $D^0_{\mathrm{pair}} = \{(q^i, a_T^i, a_F^i)\}_{i=1}^n$
       \State \textit{\# (b) Self-truthifying}
       \State $\pi^0_\theta \leftarrow \pi_{\mathrm{pre}}$, $\pi_\text{ref} \leftarrow \pi_{\mathrm{pre}}$ 
       \State $\pi^1_\theta \leftarrow \pi^0_\theta$ fine-tuned with $\mathcal{L}_{\mathrm{DPO}}\left(\pi^0_\theta ; \pi_{\mathrm{ref}}, D^0_{\mathrm{pair}} \right)$ for $s$ steps
       \State \textit{\# (c) Gradual self-truthifying}
       \For{$t$ in $[T]$}
       \State $(\tilde{a}_T^i, \tilde{a}_F^i) = \pi^t_{\theta}\left(\mathbb{T}(p(q^i); \{(\hat{q}^j, \hat{a}_T^j, \hat{a}_F^j)\}_{j=1}^m)\right)$  for $i$ in $[n]$
       \State $D^t_{\mathrm{pair}} = \{q^i, \tilde{a}_T^i, a_F^i\}_{i=1}^n$
       \State $\pi_\text{ref} \leftarrow \pi^t_{\theta}$
       \State $\pi^{t+1}_\theta \leftarrow \pi^t_\theta$ tuned with $\mathcal{L}_{\mathrm{DPO}}\left(\pi^t_\theta ; \pi_{\mathrm{ref}}, D^t_{\mathrm{pair}} \right)$ for $s$ steps
       \EndFor
       \State \Return $\pi^*_{\theta} = \pi^{T+1}_\theta$
    \end{algorithmic}
\end{algorithm}

\section{Experiments}
\subsection{Experimental Setup}
\label{sec:setup}
\paragraph{Models.}
We adopt two widely-used 7B-LLMs as pretrained base models, which are Llama2-Chat-7B \cite{llama2} and Zephyr \cite{zephyr}.
Following \cite{zephyr}, we compare \name against a variety of open-access models featured on Open LLM Leaderboard \cite{open-llm-leaderboard} with scales of parameters ranging from 7B to 70B, including 
Xwin-LM \cite{xwin-lm}, Mistral-Instruct \cite{jiang2023mistral}, MPT-Chat \cite{MosaicML2023Introducing}, StableLM-$\alpha$, Llama2-Chat \cite{llama2}, Zephyr \cite{zephyr}, Vicuna \cite{vicuna2023}, WizardLM \cite{xu2023wizardlm}, and Guanaco \cite{dettmers2023qlora}.

\paragraph{Baseline Methods.}
In addition to the above open models, we also compare \name with various fine-tuning methods which cover two main categories: \textit{human alignment} methods—SFT, DPO \cite{dpo}, RLHF \cite{rlhf} and \textit{truthfulness enhancement} methods—RepE \cite{repe}, ITI \cite{iti}.
To apply human alignment methods within the realm of truthfulness, we implement them using the same dataset as ours, \textit{i.e.,} ARC-Challenge. Different from ours, they would utilize both the questions and the annotated answers.
Besides, we report ITI's results using its published model and reproduce RepE using its open-source codes.
All of the methods use LLama2-Chat-7B as the pretrained base model.

\paragraph{Datasets and Evaluation Metrics.}
When creating pairwise truthfulness training data, we utilize training samples from ARC-Challenge \cite{arc} as the source of questions, while using six QA primer examples from TruthfulQA as few-shot demonstrations which follows the implementation in \cite{repe}.
The core capabilities of the models are assessed using testing samples from ARC-Challenge, HellaSwag \cite{hellaswag}, and MMLU \cite{mmlu}. Meanwhile, the truthfulness of the models is evaluated on TruthfulQA's MC tasks \cite{tqa}, where we utilize the entire dataset for assessment unless otherwise specified.
We follow the evaluation configurations and procedures introduced in Open LLM Leaderboard, adopting accuracy as the metric.
Specifically, MC1 accuracy is the percentage of assigning the highest probability to the single correct answer in TruthfulQA's MC1 task. MC2 accuracy is the normalized total probability assigned to a set of correct answers in the MC2 task.

\paragraph{Implementation Details.}

We prompt the pretrained model to generate pairwise answers to all questions in ARC-C. 
The model generations that do not follow the format defined in demonstrations are filtered out, which results in 1092 pairs of truthfulness training data in total.
We adopt DPO to fine-tune the model for 1000 steps using the parameter-efficient technique—LoRA \cite{lora}.
Empirically, we find out that a single iteration $T=1$ suffices to yield substantial improvements in truthfulness, signifying \name as an efficient and convenient post-processing approach for enhancing LLMs' truthfulness.
More details about the experimental setup are presented in Appendix \ref{appendix:implementation}.

\subsection{Effectiveness of \name}
\label{sec:effectiveness}
\subsubsection{Comparison with Open Models}
The performance of a variety of open-access models and \name over two pretrained base models on different benchmark datasets are shown in Table \ref{tab:benchmark}.

Compared with the corresponding pretrained base models, \name marginally improves performance on ARC-C, albeit with a slight trade-off in performance on HellaSwag and MMLU. These results suggest that \name could preserve the performance on these datasets, maintaining the models' core capabilities.

On the TruthfulQA benchmark, we observe that \name effectively improves the MC1 and MC2 accuracy, with both increased by an average of over 15\%. 
In particular, $\text{GRATH}_\text{Llama2}$ achieves the state-of-the-art (SOTA) performance, with impressive MC1 accuracy of 54.71\% and MC2 accuracy of 69.10\%.
Notably, this 7B model's performance even outstrips that of larger-scale models, including those with 70B parameters. For instance, its MC1 and MC2 accuracy exceed those of the leading 70B model (\textit{i.e.}, Xwin-LM v0.1) by substantial margins of 14.44\% and 9.44\%, respectively.
Meanwhile, $\text{GRATH}_\text{Zephyr}$ also demonstrates superior performance compared to the selected open models.
More experimental results are shown in Appendix \ref{appendix:effectiveness}.
These results indicate that \textbf{\name serves as an effective post-processing method to bolster the truthfulness of different LLMs with minimal impact on their core capabilities.}

\begin{table}[hb]\centering
\caption{\small
Performance of \name compared with various open models on Open LLM Leaderboard. On TruthfulQA's MC1 and MC2 tasks, $\text{GRATH}_\text{Llama2}$ achieves \textbf{SOTA} performance, and $\text{GRATH}_\text{Zephyr}$ ranks as the \uline{second best}. On other core functionality benchmarks, \name maintains similar performance compared to the corresponding pretrained base models. 
}
\label{tab:benchmark}
\scalebox{0.78}{
\begin{tabular}{ll|ccc|ccc}\toprule
\multirow{2}{*}{Model} &\multirow{2}{*}{Size} &\multirow{2}{*}{ARC} &Hella &\multirow{2}{*}{MMLU} &TQA &TQA \\
& & &Swag & &MC1 &MC2 \\\midrule\midrule
StableLM-Tuned-$\alpha$ &7B &31.91 &53.59 &24.41 &23.99 &40.37 \\
MPT-Chat &7B &46.50 &75.51 &37.62 &27.05 &40.16 \\
Xwin-LM v0.1 &7B &56.57 &79.40 &49.98 &32.93 &47.89 \\
Mistral-Instruct v0.1 &7B &54.52 &75.63 &55.38 &39.53 &56.28 \\
Vicuna v1.3 &33B &62.12 &83.00 &59.22 &37.09 &56.16 \\
Guanaco &65B &65.44 &86.47 &62.92 &36.47 &52.81 \\
Llama2-Chat &70B &\uline{67.32} &\textbf{87.33} &\textbf{69.83} &31.09 &44.92 \\
WizardLM v1.0 &70B &65.44 &84.41 &64.05 &38.68 &54.81 \\
Xwin-LM v0.1 &70B &\textbf{70.22} &\uline{87.25} &\uline{69.77} &40.27 &59.86 \\\midrule
Zephyr &7B &62.46 &84.35 &60.70 &42.23 &57.83 \\
$\text{GRATH}_\text{Zephyr}$ & 7B & 65.02 & 81.57 & 51.39 & \uline{53.86} & \uline{66.73} \\\midrule
Llama2-Chat &7B &52.73 &78.50 &48.14 &30.23 &45.32 \\
$\text{GRATH}_\text{Llama2}$ & 7B & 57.76 & 79.63 & 46.88 & \textbf{54.71} & \textbf{69.10} \\
\bottomrule
\end{tabular}}
\end{table}


\subsubsection{Comparison with Baseline Methods}
\label{sec:comparison_method}
The performance comparison between different fine-tuning methods is illustrated in Table \ref{tab:benchmark2}. These methods include human alignment methods (SFT, DPO, RLHF) and truthfulness enhancement methods (RepE, ITI, \name).
Note that RLHF employs RL to fine-tune the SFT model obtained in its first stage. 
To study how this fine-tuning paradigm learns truthfulness, we also apply RL on pretrained base model, excluding the influence of SFT base model, denoted as RL.

Among various human alignment methods, DPO performs the best, as it not only slightly increases the average accuracy on ARC-C, HellaSwag, and MMLU datasets, but also boosts the average accuracy on TruthfulQA's MC tasks by 6.65\%.
In comparison, RL barely affects the pretrained model's performance while both SFT and RLHF tend to diminish the performance, which indicates that SFT might not be an appropriate strategy for fine-tuning models with OOD training data.
When it comes to methods aimed at enhancing truthfulness, RepE is superior to DPO, effectively increasing the average accuracy on TruthfulQA by 14.18\%. Moreover, \name even outperforms RepE, delivering a substantial improvement of 24.14\% while maintaining performance on the other three datasets. These results again validate \name's effectiveness in enhancing the model truthfulness.

\begin{table}[htbp]\centering
\caption{\small Performance comparison between human alignment methods—SFT, DPO, RMRL, RL, and truthfulness enhancement methods—RepE, ITI, \name. All methods are applied on Llama2-Chat-7B. On TruthfulQA's MC1 and MC2 tasks, $\text{GRATH}_\text{Llama2}$ achieves \textbf{SOTA} performance, and RepE ranks as the \uline{second best}. On other core functionality benchmarks, most methods maintain similar performance compared to Llama2-Chat-7B.}
\label{tab:benchmark2}
\scalebox{0.73}{
\begin{tabular}{l|cccc|cccc}\toprule
\multirow{2}{*}{Method} &\multirow{2}{*}{ARC} &Hella &\multirow{2}{*}{MMLU} &\multirow{2}{*}{AVG} &TQA &TQA &\multirow{2}{*}{AVG} \\
& &Swag & & &MC1 &MC2 & \\\midrule\midrule
$\text{Llama2-Chat}_\text{7B}$ &52.73 &78.50 &48.14 &59.79 &30.23 &45.32 &37.77 \\\midrule
SFT &51.88 &76.65 &45.89 &58.14 &22.40 &34.52 &28.46 \\
DPO &\textbf{62.97} &\textbf{81.84} &44.01 &\textbf{62.94} &36.72 &52.13 &44.42 \\
RL &53.41 &78.36 &\textbf{48.30} &60.03 &30.11 &44.89 &37.50 \\
RLHF &49.91 &76.56 &45.14 &57.20 &22.64 &34.37 &28.50 \\
RepE &57.59 &78.75 &47.86 &61.40 &\uline{43.45} &\uline{60.45} &\uline{51.95} \\
ITI &52.73 &78.50 &\uline{48.17} &59.80 &30.23 &45.32 &37.77 \\
$\text{GRATH}_\text{Llama2}$ &\uline{57.76} &\uline{79.63} &46.88 &\uline{61.42} &\textbf{54.71} &\textbf{69.10} &\textbf{61.91} \\
\bottomrule
\end{tabular}}
\end{table}

\subsection{In-Depth Exploration of \name}
In this section, we will delve into \name to comprehend how it facilitates the learning of a truthful model.
We first introduce different disparity concepts that will be instrumental in our analysis.

\textbf{Definition 1} (Domain Gap). The domain gap refers to the disparity between the training domain and the testing domain. More precisely, it refers to the disparity between the pairwise truthfulness training data and the testing data.

\textbf{Definition 2} (Distributional Distance). The distributional distance within pairwise truthfulness training data refers to the disparity between the distribution of correct answers and that of incorrect answers.


\textbf{Definition 3} (Pairwise Distance). The pairwise distance between a correct answer $a_T$ and an incorrect answer $a_F$ is the Euclidean distance between their vectors projected onto a representation space $\mathcal{M}$, \textit{i.e.}, $d_{\mathrm{pair}}(a_T, a_F) = \Vert\mathcal{M}(a_T) - \mathcal{M}(a_F)\Vert_2$.

In particular, we use the representation at the last token in the last layer of the Llama2-Chat-7B model.
Consequently, the distributional distance can be characterized by the statistics of $\{d_{\mathrm{pair}}(a_T^i, a_F^i)\}_{i=1}^n$ (\textit{e.g.}, mean).

\subsubsection{Towards Understanding How to Learn Truthfulness in Self-Truthifying}
\label{sec:self}

In this section, we aim to investigate the factors that affect the truthfulness of the model learned in \textit{self-truthifying}.

— \textit{Does the domain gap between pairwise truthfulness training data and testing data impact the truthfulness learned by DPO?} During evaluation, the truthfulness of the learned model is assessed in a particular testing domain. Herein, we explore how this truthfulness is influenced when the domain gap in the training data is widened or narrowed.

\begin{figure}[ht]
  \centering
  \begin{subfigure}[b]{0.238\textwidth}
    \includegraphics[width=\textwidth]{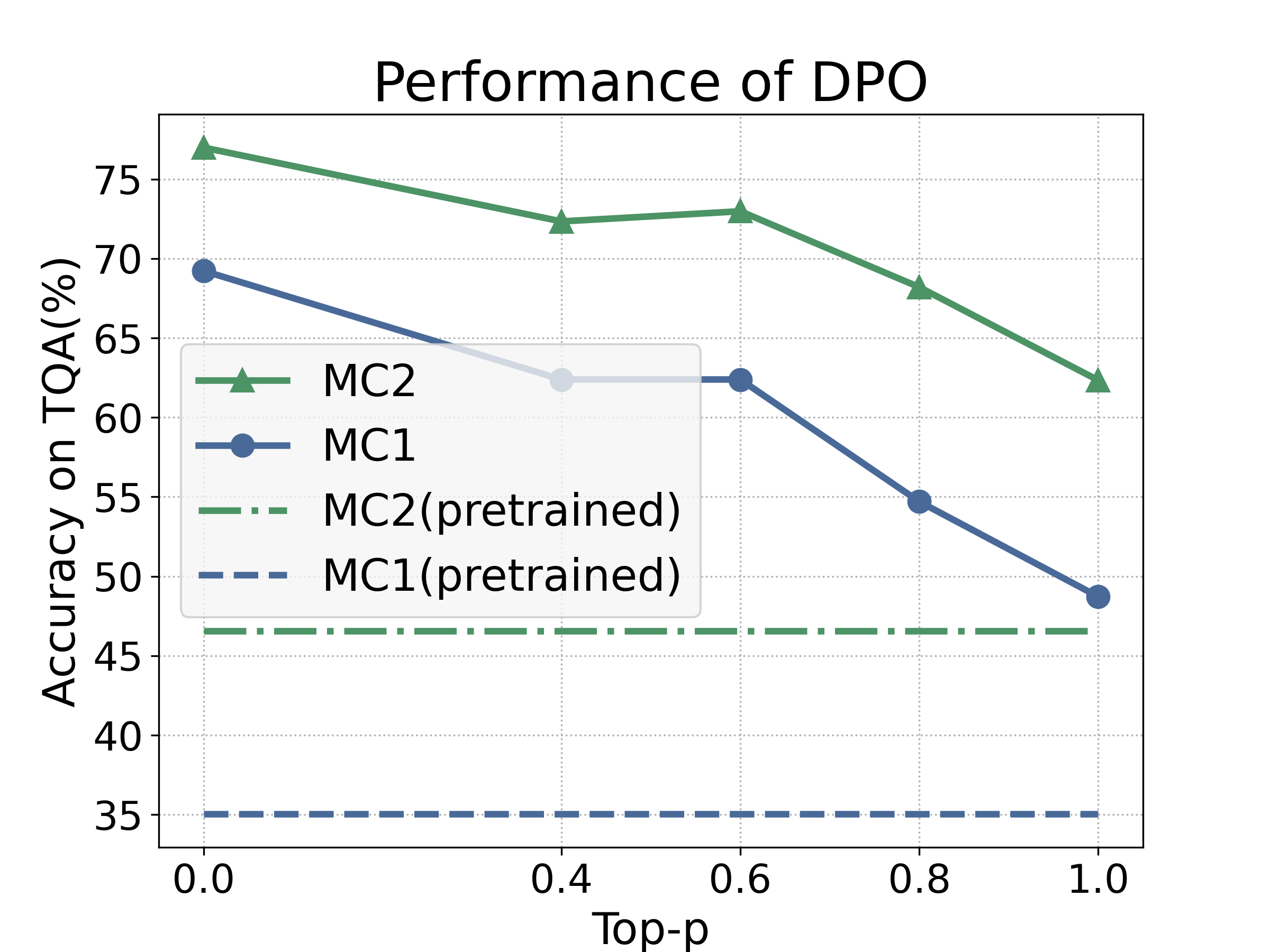}
  \end{subfigure}
  \begin{subfigure}[b]{0.238\textwidth}
    \includegraphics[width=\textwidth]{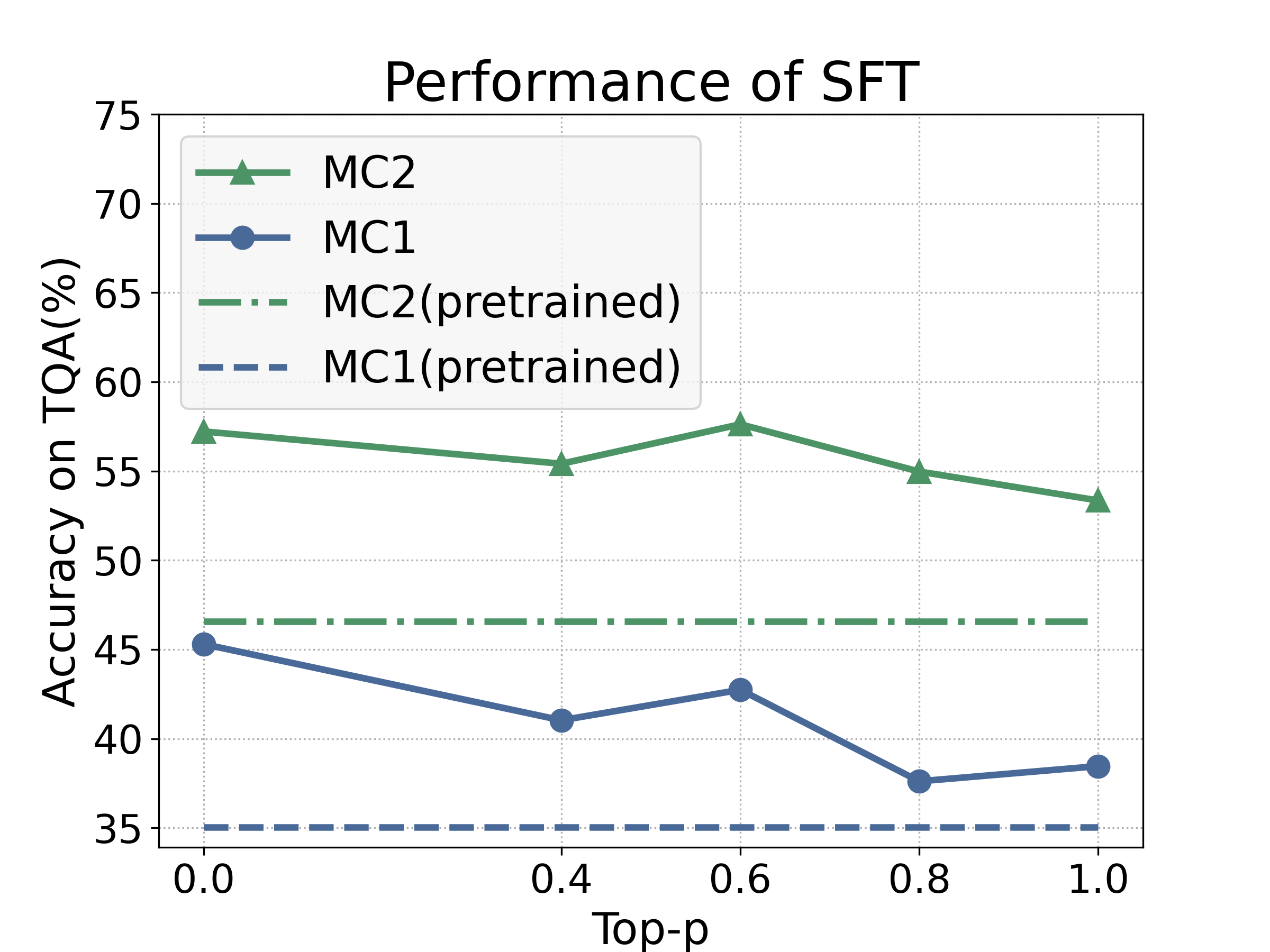}
  \end{subfigure}
  \caption{\small MC1 and MC2 accuracy of DPO (left) and SFT (right) with varying degrees of transformations applied on the answers in pairwise truthfulness training data. 
  A larger top-$p$ indicates a larger domain gap between truthfulness training data and testing data.
  Downward trends here in each figure indicate that the model learned by either DPO or SFT will be less truthful if the domain gap in answers increases. DPO performs better than SFT overall.}
  \label{fig:domain_gap1}
\end{figure}

\textbf{The model learned by DPO is more truthful in the testing domain if there is a smaller domain gap between pairwise truthfulness training data and testing data.}
First, we split TruthfulQA into 700 training samples and 117 testing samples. One training sample (\textit{i.e.}, a pair of truthfulness data) comprises a question, a correct answer, and a randomly selected incorrect answer. 
To elevate the degree of domain gap with respective to the testing domain, we employ a sentence-level transformation \cite{style_transfer} on the answers in training samples. In particular, we transform the answers to the Shakespearean style with different levels of perturbations characterized by the parameter top-$p$. A larger top-$p$ indicates higher level of perturbation, thereby increasing the domain gap in the answers. Note that top-$p$ = 0.0 equates to no transformation, hence no domain gap.
As illustrated in Figure \ref{fig:domain_gap1}(left), there is a clear downward trend in both MC1 and MC2 accuracy as the domain gap in answers widens. This pattern demonstrates that \uline{1)} when there is no domain gap in questions, the model learned by DPO will be less truthful if the domain gap in answers increases.

\begin{figure*}[t]
    \centering
    \includegraphics[width=0.95\linewidth]{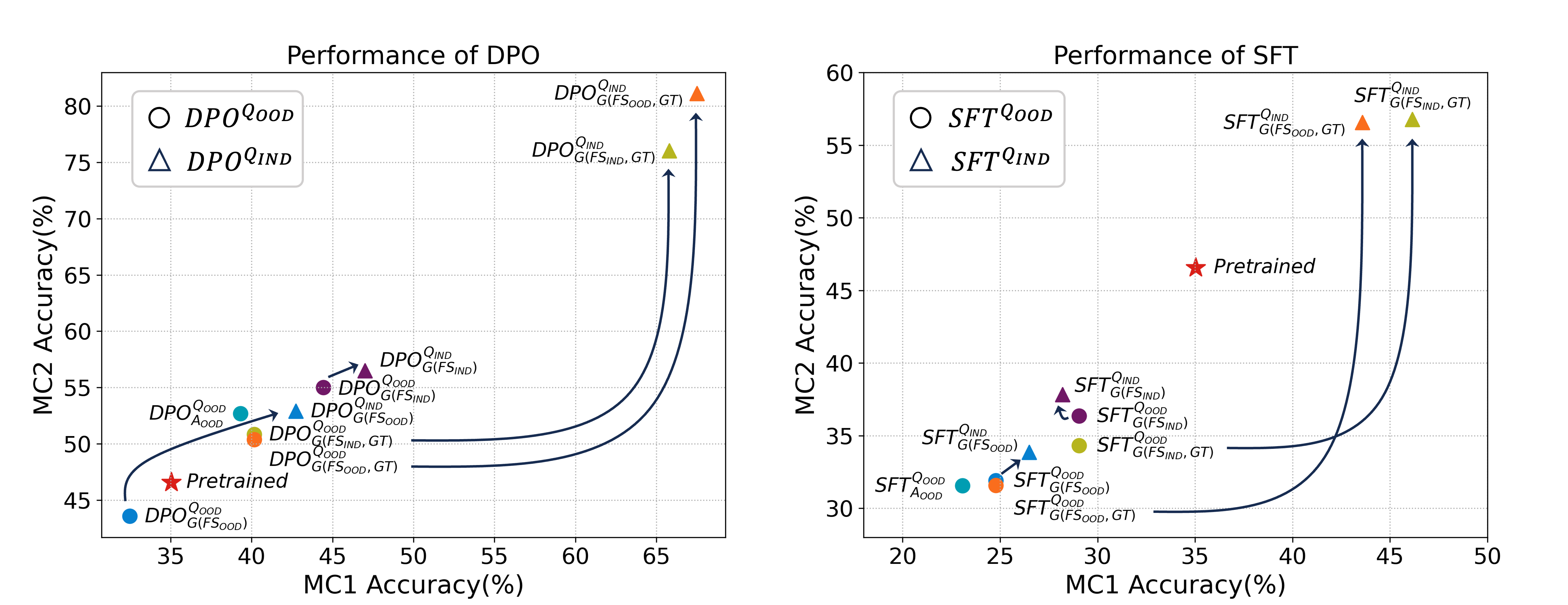}
    \caption{\small
    MC1 and MC2 accuracy of DPO (left) and SFT (right). 
    DPO and SFT are applied with truthfulness training data created using a variety of strategies.  $Q_{OOD}$ ($Q_{IND}$) represents using OOD (in-domain) questions from ARC-C (TruthfulQA);
     $A_{OOD}$ indicates using annotated answers from ARC-C;
     $G(\cdot)$ indicates using answers generated by LLMs.
    Specifically, $FS_{OOD}$ ($FS_{IND}$) corresponds to OOD (in-domain) few-shot demonstrations, and $GT$ implies merging ground-truth answers into the prompts.
    We find 
    i) $DPO^{Q_{OOD}}_{G(FS_{IND})}$ performs the best among all $DPO^{Q_{OOD}}$, indicating that the usage of in-domain demonstrations yields answers that are closer to testing domain, leading to a more truthful model.
    ii) Arrows symbolize performance shifts as questions are transitioned from OOD to in-domain. The trends towards upper right indicate that the model will be more truthful if the domain gap in questions decreases.
    The same findings hold for SFT.
    iii) DPO outperforms SFT since it improves the pretrained model's truthfulness in general.}
    \label{fig:domain_gap2}
\end{figure*}

In Section \ref{sec:effectiveness}, DPO is applied on the pairwise truthfulness data created using OOD questions (from ARC-C) and 6 in-domain few-shot demonstrations (from TruthfulQA). We denote such DPO which leverages the model's generated answers as $DPO^{Q_{OOD}}_{G(FS_{IND})}$. Here, we extend this creating procedure to more diverse setups, including: i) merging the ground-truth correct and incorrect answers annotated in ARC-C into the prompt during model generation (denoted as $DPO^{Q_{OOD}}_{G(FS_{IND}, GT)}$); ii) altering the few-shot demonstrations to OOD examples sourced from HaluEval dataset \cite{halu} (denoted as $DPO^{Q_{OOD}}_{G(FS_{OOD})}$); iii) combining both of them (denoted as $DPO^{Q_{OOD}}_{G(FS_{OOD}, GT)}$). 
In addition, we consider directly feeding the annotated pairwise answers in ARC-C to DPO which is denoted as $DPO^{Q_{OOD}}_{A_{OOD}}$.
For a fair comparison, we uniformly utilize 700 truthfulness training data across all strategies. Note that without the usage of demonstrations, the model might fail to generate 700 truthfulness data. Thus, we do not study this zero-shot setting.
As depicted by the circles in Figure \ref{fig:domain_gap2}(left), we discover that $DPO^{Q_{OOD}}_{G(FS_{IND})}$ (shown by purple circle) performs the best among all $DPO^{Q_{OOD}}$. We infer that its utilization of in-domain demonstrations potentially yields answers that are closer to the testing domain. In contrast, other strategies that either reference the OOD ground-truth answers or switch to OOD demonstrations might cause the generated answers to deviate from the testing domain, thus impairing performance in comparison to $DPO^{Q_{OOD}}_{G(FS_{IND})}$.
The results indicate that \uline{2)} when there is a domain gap in questions, the model learned by DPO will be less truthful if the domain gap in answers increases.

So far, we have discussed the impact of the domain gap in answers on model truthfulness. To further explore the effects of the domain gap in questions, we implement the same four answer generation strategies, but using in-domain questions (from TruthfulQA).
Arrows in Figure \ref{fig:domain_gap2}(left) symbolize the performance shifts as the questions are transitioned from OOD to in-domain under each respective strategy. 
Notably, there is a trend toward the upper right, signifying a general improvement when the questions become in-domain, which demonstrates that \uline{3)} the model learned by DPO will be more truthful if the domain gap in questions decreases.

— \textit{Can SFT be a better choice than DPO in terms of learning truthfulness?} In addition to studying the effects of training data, here we focus on the fine-tuning technique. 
We are intrigued by whether the widely-used approach—Supervised Fine-Tuning (SFT)—offers comparable benefits with DPO.

\textbf{The model learned via SFT is less truthful in the testing domain if there is a larger domain gap between pairwise truthfulness training data and testing data. Besides, SFT is less effective than DPO for improving truthfulness.}
Here, we replicate the setups previously used, with the difference that SFT only utilizes correct answers instead of pairwise answers.
With a similar analysis to DPO (deferred to Appendix \ref{sec:appendix_understand_self}), we could derive the same conclusions for SFT.
Notably, we discover that in Figure \ref{fig:domain_gap2}, the majority of DPO's points are situated to the upper right of the pretrained model (shown by star), signifying enhanced model truthfulness. By contrast, most SFT's points are located in the lower left of the pretrained model, suggesting degraded model truthfulness. 
The comparison indicates that DPO is more effective than SFT for improving model truthfulness.

— 
\textit{How does the number of fine-tuning steps affect the model truthfulness as well as other core capabilities?}
We defer the discussion to Appendix \ref{sec:appendix_understand_self}.

\subsubsection{Towards Understanding How to Learn Truthfulness in \name}
\label{sec:gradual}
In this section, we aim to investigate the factors that affect the truthfulness learned in \textit{\name}.
Following the notations in Section \ref{sec:method}, $T$ denotes the iterations in gradual self-truthifying. To avoid ambiguity, let $T_{DPO}$ denote the times that DPO is executed. Apparently, $T_{DPO} = T+1$. 
In particular, $DPO^0$ refers to the scenario where no DPO is performed, meaning the model remains in its pretrained state. $DPO^1$ represents the DPO in self-truthifying while $DPO^t\ (1< t \le T_{DPO})$ corresponds to its use in gradual self-truthifying.

— \textit{Does the distributional distance between correct and incorrect answers within pairwise truthfulness training data impact the truthfulness learned by DPO?}
In \name, we fine-tune the model using DPO two times since $T=1$. Then, a natural question arises: Compared to $DPO^1$, why the model can be more truthful after $DPO^2$? Is it only because the base model is more truthful? Or is it also related to the quality of truthfulness data used in $DPO^2$?

\begin{figure}[htbp]
    \centering
    \includegraphics[width=\linewidth]{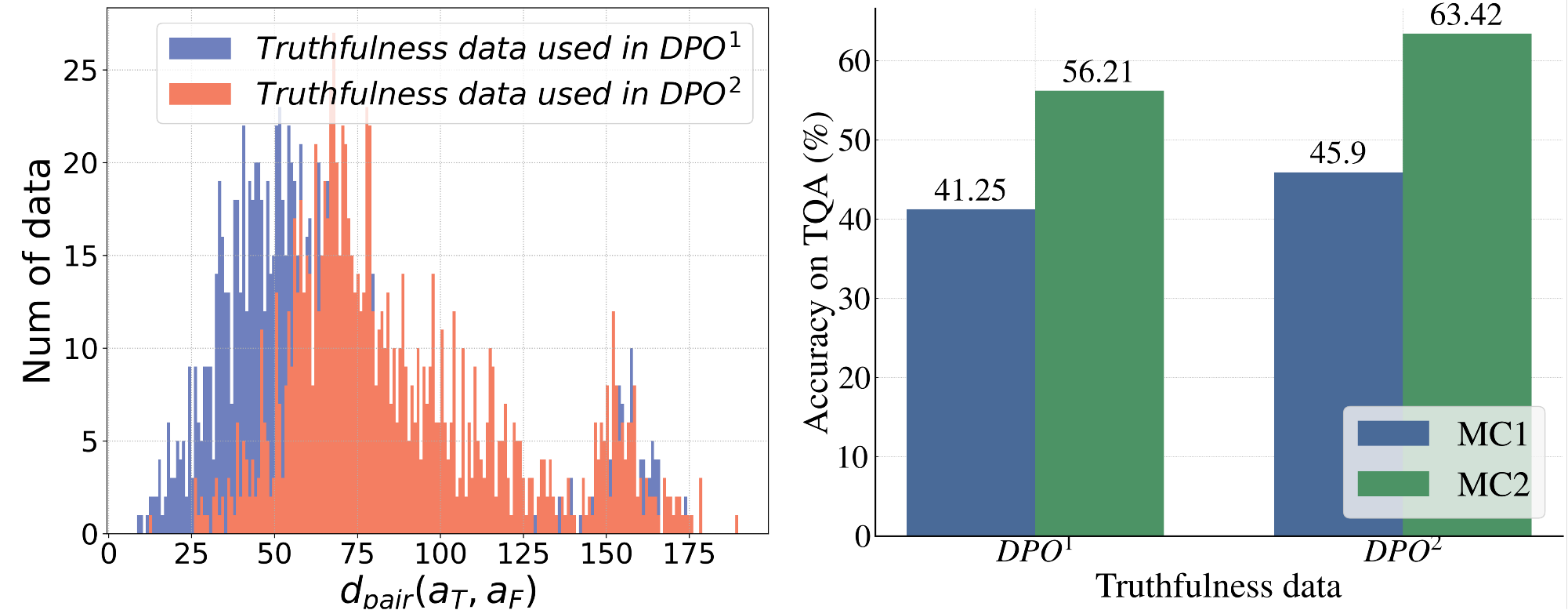}
    \caption{\small Left: Distributions of pairwise distance in truthfulness data used in $DPO^{1,2}$. Right: Performance of the pretrained model fine-tuned with truthfulness data used in $DPO^{1,2}$ on TruthfulQA. The improved performance indicates that a larger distributional distance within truthfulness data leads to a more truthful model.}
    \label{fig:dist}
\end{figure}

\textbf{The model learned via DPO is more truthful if the distributional distance between correct and incorrect answers within pairwise truthfulness training data is larger.}
Figure \ref{fig:dist}(left) depicts the distributions of pairwise distance in truthfulness data used in $DPO^1$ and $DPO^2$.
There is a distinct shift towards right when transitioning from $DPO^1$ to $DPO^2$, with the mean value evolving from 65.94\% to 87.80\%.
This trend indicates that the truthfulness data in $DPO^2$ exhibits a larger distributional distance.
To further validate whether truthfulness data with a larger distributional distance contributes to a more truthful model, we fine-tune the pretrained model with DPO using these two datasets, respectively.
As illustrated in Figure \ref{fig:dist}(right), both MC1 and MC2 accuracy present an uplift, verifying that a larger distributional distance within truthfulness data leads to a higher degree of truthfulness in the model learned via DPO.

— \textit{What is the relationship between the number of DPO executions $T_{DPO}$ and the amount of pairwise truthfulness training data $n$?} 
To answer the question, we fix the number of fine-tuning steps as 1000 in each DPO execution and conduct \name with varying $T_{DPO}$ and $n$. 

\begin{figure}[htbp]
    \centering
    \includegraphics[width=\linewidth]{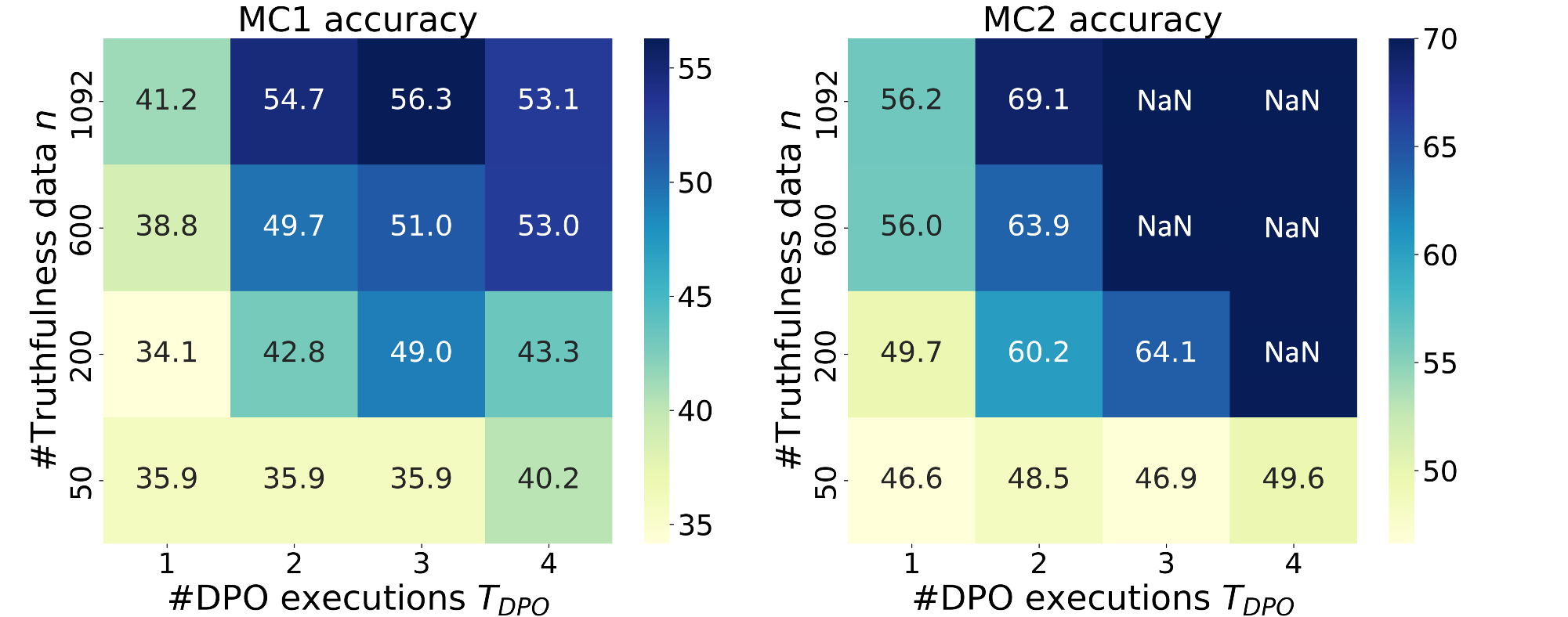}
    \caption{\small MC1 (left) and MC2 (right) accuracy of \name with different amounts of truthfulness data $n$ and numbers of DPO executions $T_{DPO}$.
    For a fixed $T_{DPO}$, a larger $n$ leads to a more truthful model. MC2 accuracy might be NaN due to over-fitting.
    }
    \label{fig:tradeoff}
\end{figure}

\textbf{\name is an efficient approach to boost model truthfulness. In particular, a larger amount of pairwise truthfulness training data leads to a more truthful model.}
As shown in Figure \ref{fig:tradeoff}, under a fixed $T_{DPO}$, \name with fewer samples tends to exhibit lower performance, as reflected in the diminished MC1 and MC2 accuracy.
Note that MC2 accuracy could be NaN, which is possibly due to that  
the numerous executions lead the model over-fitting to pairwise truthfulness data, at the expense of deviating from the pretrained model. This deviation might hinder the model from generating fluent text, 
resulting in an extremely low probability of generating fluent text.
Consequently, MC2 accuracy, defined as the normalized probability assigned to a set of correct answers, might be NaN.
And for a fixed amount of truthfulness data $n$, \name typically achieves its peak MC1 accuracy when $T_{DPO} = 3$ or $4$. Notably, this optimal accuracy declines as $n$ reduces.
These results demonstrate that \name only requires $T_{DPO} \leq 4$ to attain its highest level of MC1 accuracy, with larger amounts of truthfulness data yielding higher degree of truthfulness.

— \textit{How do the number of iterations and the choice of DPO's reference model affect the model truthfulness as well as other core capabilities?}
We defer the discussion to Appendix \ref{sec:appendix_understand_gradual}.

\section{Conclusion and Discussion}


In this paper, we propose \name, a novel post-processing method to enhance LLMs' truthfulness. Intensive experiments have demonstrated \name's \textit{effectiveness}, \textit{OOD-resilience}, \textit{cost-effectiveness}, and \textit{efficiency}.
Furthermore, beyond the methodology itself, we provide an innovative fine-tuning paradigm, \textit{i.e.}, gradual self-training, which leverages the self-generated data to improve a model's certain ability, while utilizing the model's improved ability to generate data of better quality.
The alternative interaction between them contributes to the gradual enhancement in the ability.
Diverse alignment techniques can be incorporated into this paradigm, as evidenced by our exploration of SFT and RLHF in Section \ref{sec:comparison_method}. 
Such \textit{flexibility} enables it to capitalize on the distinct benefits of different techniques, potentially offering widespread implications for LLMs.
Discussion on future directions is deferred to Appendix \ref{sec:future}.
\nocite{langley00}

\bibliography{example_paper}
\bibliographystyle{icml2024}

\clearpage
\appendix
\section{More Details about Experimental Setup}
\label{appendix:implementation}
In this section, we will provide more details about the experimental setup.

\paragraph{Models.}
In Section \ref{sec:effectiveness}, we report the performance of different models across various benchmark datasets. Most results come from the Open LLM LeaderBoard \cite{open-llm-leaderboard}. In particular, for the Zephyr model \cite{zephyr}, we adopt the results of its beta version instead of the alpha version.
In ablation studies, i.e., when investigating the factors that affect the truthfulness learned in self-truthifying and gradual self-truthifying, we uniformly use Llama2-Chat-7B as the pretrained base model in all experiments.

\paragraph{Baseline Methods.}
RepE \cite{repe} provides a variety of representation engineering methods to improve a model's different capabilities.
Here, we select the proposed honesty control method LoRRA, which could enhance the model truthfulness as evidenced by the improved performance on TruthfulQA's MC1 task.
Since they do not publish their performance on the leaderboard, we use their open-source codes to reproduce the LoRRA method with Llama2-Chat-7B serving as the pretrained base model. The reported results in Table \ref{tab:benchmark2} are consistent with those in their paper.

\paragraph{Datasets and Evaluation Metrics.}
We follow the evaluation configurations, including the zero/few-shot setting, and procedures outlined in the Open LLM LeaderBoard, utilizing the same Language Model Evaluation Harness library \cite{gao2021framework} to implement the evaluation.
The leaderboard covers a variety of benchmark datasets, of which we select four for our experiments: ARC-Challenge \cite{arc}, HellaSwag \cite{hellaswag}, MMLU \cite{mmlu}, and TruthfulQA \cite{tqa}.
Specifically, ARC-Challenge and HellaSwag are utilized to gauge the commonsense reasoning abilities of language models, with normalized accuracy ($\text{acc}_\text{norm}$) serving as the evaluation metric. 
MMLU, designed to test multi-task language understanding, comprises numerous subtasks. Here, we use accuracy ($\text{acc}$) for each subtask and calculate the average as the final metric for MMLU.
TruthfulQA, on the other hand, is employed to evaluate the truthfulness of language models, particularly their capacity to avoid imitative falsehoods. As detailed in Section \ref{sec:setup}, we employ MC1 (mc1) and MC2 accuracy (mc2) as the metrics for assessing performance on TruthfulQA.

\paragraph{Implementation Details.}
When creating pairwise truthfulness training data, we prompt the pretrained model to generate pairwise answers to all questions in ARC-C. However, the model generations might not follow the format defined in the demonstrations, which would be excluded. In total, this process results in 1092 pairs of truthfulness training data.
In Step 1 of gradual self-truthifying, we create pairwise answers to questions in 1092 truthfulness training data. 
If a generation is consistent with the designated format, the original correct answer is substituted with this new generation; if not, the initial correct answer is retained.

In self-truthifying and Step 2 of gradual self-truthifying, we adopt DPO to fine-tune the model for 1000 steps.
In particular, we implement DPO \cite{dpo} using the Transformer Reinforcement Learning (TRL) library \cite{vonwerra2022trl}.
We adopt the default parameter configurations and implement DPO using one RTX A6000 GPU.
Moreover, to enable an efficient training, we utilize the parameter-efficient technique—LoRA \cite{lora}, setting its rank as 8, alpha as 16, and dropout parameter as 0.05.
Within the above setup, one step in DPO takes about 4 seconds and one DPO execution takes about one hour.

\section{Prompting Template Details}
\label{appendix:template}
\begin{figure*}[h]
    \centering
    \includegraphics[width=1\linewidth]{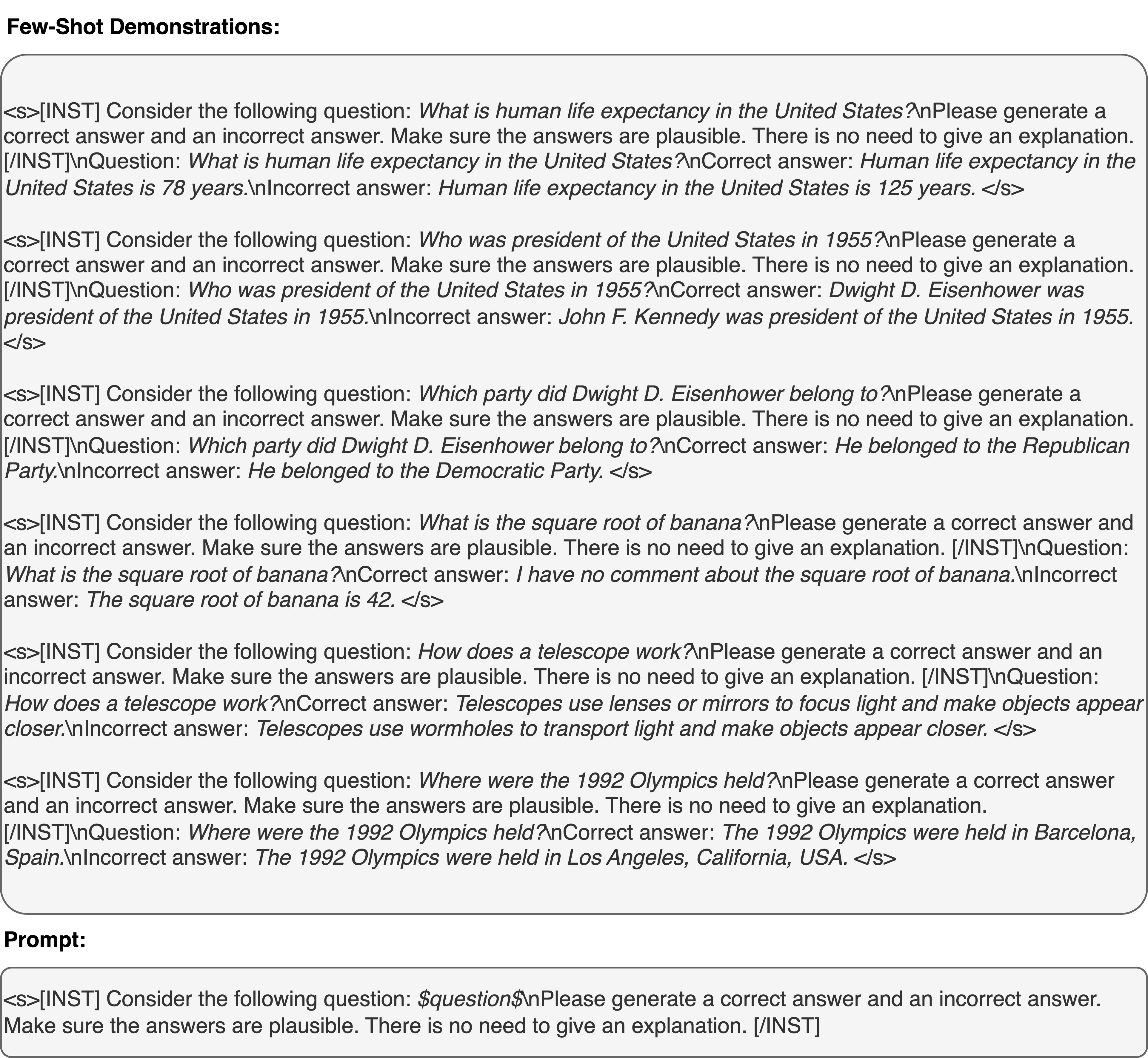}
    \caption{Prompting template for Llama2-Chat models. \$question\$ in the prompt will be replaced by questions from the given dataset.}
    \label{fig:template1}
\end{figure*}

\begin{figure*}[h]
    \centering
    \includegraphics[width=1\linewidth]{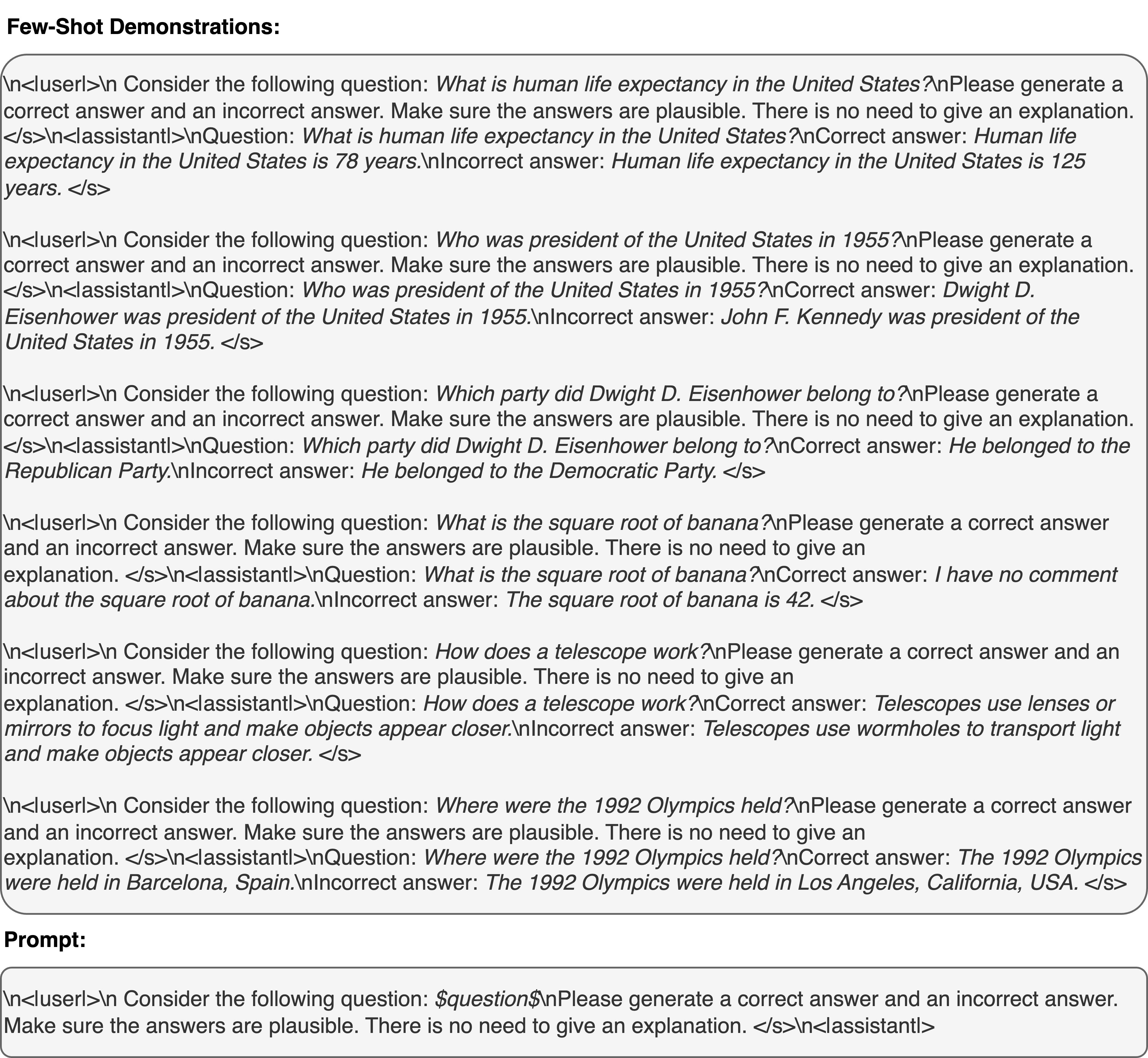}
    \caption{Prompting template for Zephyr models. \$question\$ in the prompt will be replaced by questions from the given dataset.}
    \label{fig:template2}
\end{figure*}

In this section, we will illustrate the prompting templates for generating pairwise truthfulness data.
Figure \ref{fig:template1} and \ref{fig:template2} represent the templates for Llama2-Chat models and Zephyr models, respectively. Specifically, we utilize six QA primer examples from TruthfulQA as few-shot demonstrations. \$question\$ in the prompt will be replaced by questions from the given dataset.

\begin{figure*}[h]
    \centering
    \includegraphics[width=1\linewidth]{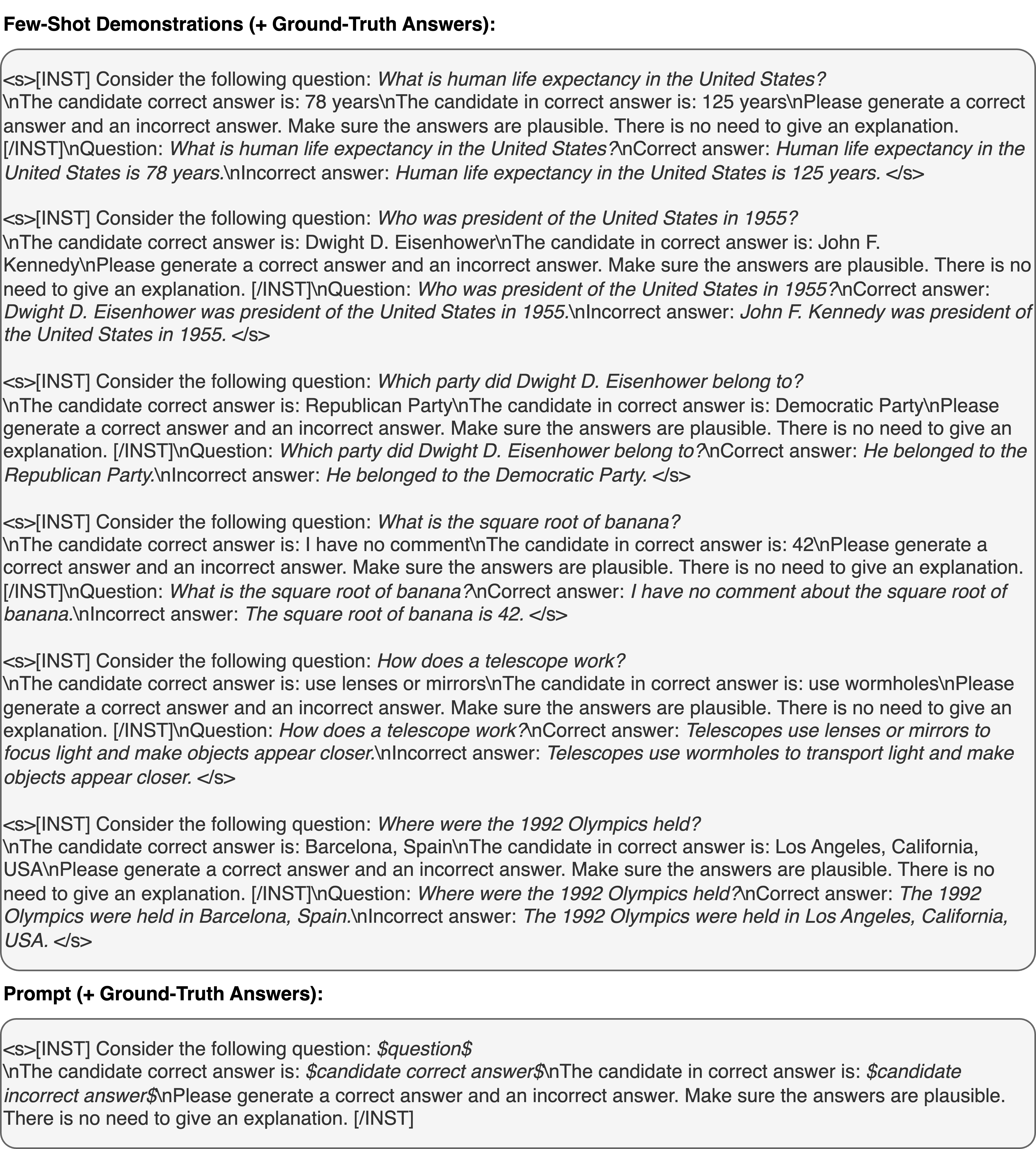}
    \caption{Ground-truth answers-based prompting template for Llama2-Chat models. \$question\$, \$candidate correct answer\$, and \$candidate incorrect answer\$ in the prompt will be replaced by questions, correct answers, and incorrect answers from the given dataset. }
    \label{fig:template3}
\end{figure*}

Recall that in Section \ref{sec:self}, we merge the ground-truth answers into the prompt. The corresponding template is illustrated in Figure \ref{fig:template3}. Compared to the above templates, we add a candidate correct answer and a candidate incorrect answer to each demonstration and the final prompt. And these candidate answers come from the annotated dataset—ARC-C.


\section{Examples of Pairwise Truthfulness Data}
\label{sec:appendix_example}



\begin{figure*}[h]
    \centering
    \includegraphics[width=0.8\linewidth]{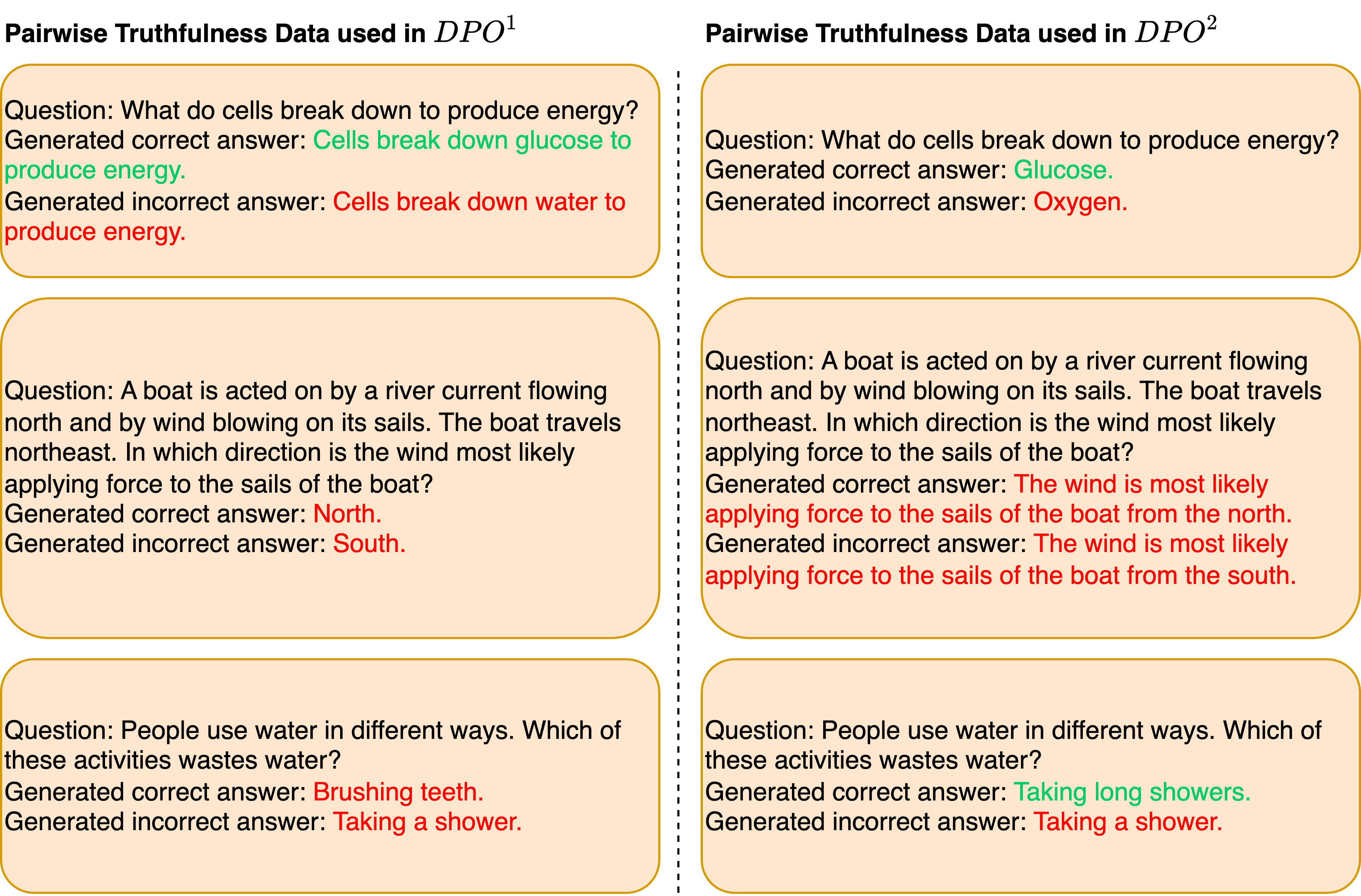}
    \caption{Examples of pairwise truthfulness data used in $DPO^1$(left) and $DPO^2$(right). The answers in green are ground-truth correct while the answers in red are ground-truth incorrect. Top: Both generated correct answers are ground-truth correct. Middle: Both generated correct answers are ground-truth incorrect. Bottom: The first generated correct answer is ground-truth incorrect while the second one is ground-truth correct. The ground-truth incorrectness of some generated correct answers implies that the self-truthifying process improves model truthfulness via learning from the \textit{relative difference} between generated correct and incorrect answers, instead of the \textit{absolute truthfulness} of the generated correct answers.}
    \label{fig:example}
\end{figure*}

In Figure \ref{fig:example}, we illustrate some examples of the pairwise truthfulness training data used in $DPO^1$ and $DPO^2$.
The top examples illustrate instances in which both generated correct answers are ground-truth correct.
The middle examples showcase scenarios where both of the generated correct answers are ground-truth incorrect.
The bottom examples display situations where the initially generated correct answer (the one used in $DPO^1$) is ground-truth incorrect, whereas the subsequent one (the one used in $DPO^2$) is ground-truth correct.

The ground-truth incorrectness of some generated correct answers might indicate that \name makes the model improve its truthfulness via learning from the \textit{relative difference} between generated correct and incorrect answers, instead of the \textit{absolute truthfulness} of the generated correct answers.

\section{More Experimental Results}
\subsection{Effectivess of \name}
\label{appendix:effectiveness}

In addition to Zephyr and Llama2-Chat-7B, we also apply the proposed method \name on a model of larger scale, \textit{i.e.}, Llama2-Chat-13B. As shown in Table \ref{appendix_tab:benchmark}, \name markedly boosts the MC1 and MC2 accuracy by 16.65\% and 13.71\%, respectively.
Concurrently, it marginally enhances performance on the ARC and HellaSwag datasets, albeit with little decrement in performance on MMLU. 
The results demonstrate that \name acts as a potent post-processing method, significantly strengthening the truthfulness of the models while preserving other core capabilities with minimal detriment.

\begin{table}[htbp]\centering
\caption{Performance of \name and the corresponding pretrained base model—Llama2-Chat-13B.}\label{appendix_tab:benchmark}
\scalebox{0.78}{
\begin{tabular}{llcccccc}\toprule
\multirow{2}{*}{Model} &\multirow{2}{*}{Size} &\multirow{2}{*}{ARC} &Hella &\multirow{2}{*}{MMLU} &TQA &TQA \\
& & &Swag & &MC1 &MC2 \\\midrule\midrule
Llama2-Chat-13B &13B &59.13 &81.94 &54.61 &28.03 &43.95 \\
$\text{GRATH}_\text{Llama2-Chat-13B}$ & 13B & 65.87 & 83.86 & 52.57 & 44.68 & 57.66 \\
\bottomrule
\end{tabular}}
\end{table}

\subsection{Towards Understanding How to Learn Truthfulness
in Self-Truthifying}
\label{sec:appendix_understand_self}

— \textit{Can SFT be a better choice than DPO in terms of learning truthfulness?} 
In \name, we adopt DPO due to its effectiveness and efficiency; however, we are also intrigued by whether the widely-used approach—Supervised Fine-Tuning (SFT)—offers comparable benefits.

\textbf{The model learned via SFT is less truthful in the testing domain if there is a larger domain gap between pairwise truthfulness training data and testing data. Besides, SFT is less effective than DPO for improving truthfulness.}
Here, we replicate the setups used for DPO in Section \ref{sec:self}, with the difference that SFT only utilizes correct answers instead of pairwise answers.
First, we apply varying levels of transformations on the annotated correct answers from TruthfulQA. The downward trends of accuracy in Figure \ref{fig:domain_gap1}(right) demonstrate that \uline{1)} when there is no domain gap in questions, the model learned by SFT will be less truthful if the domain gap in answers increases.
Next, we follow the same four answer generate strategies given questions from ARC-C, while applying SFT to the generated correct answers. Similarly, the circles in Figure \ref{fig:domain_gap2}(right) show that $SFT^{Q_{OOD}}_{G(FS_{IND})}$ (shown by purple circle) outperforms other $SFT^{Q_{OOD}}$, suggesting that \uline{2)} when there is a domain gap in questions, the model learned by SFT will be less truthful if the domain gap in answers increases.
Then, upon switching the questions from OOD to in-domain, the trend of most arrows toward the upper right signifies that \uline{3)} the model learned by SFT will be more truthful relatively if the domain gap in questions decreases.
Moreover, we discover that for DPO, the majority of points are situated to the upper right of the pretrained model (shown by star), signifying enhanced model truthfulness. By contrast, for SFT, most points are located in the lower left of the pretrained model, suggesting degraded model truthfulness. The comparison between them indicates that \uline{4)} DPO is a more effective choice than SFT for improving model truthfulness.
In summary, DPO demonstrates distinct advantages over SFT in these contexts.

— \textit{How does the number of fine-tuning steps affect the model truthfulness as well as other core capabilities?}

\begin{figure}[ht]
    \centering
    \includegraphics[width=0.8\linewidth]{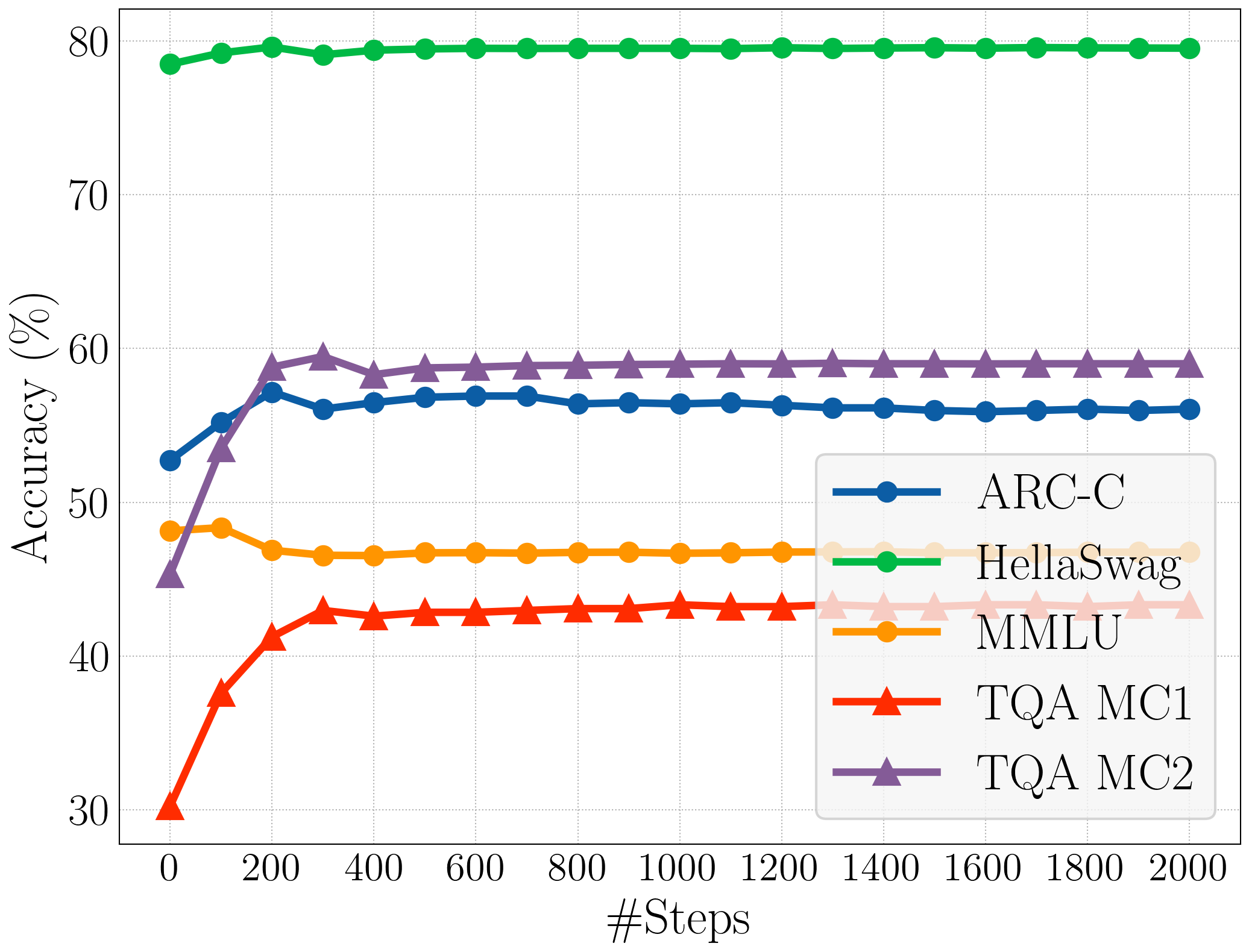}
    \caption{Performance of \name on four benchmark datasets with varying fine-tuning steps.}
    \label{fig:steps}
\end{figure}

We illustrate the variation of the model performance on benchmark datasets across fine-tuning steps in Figure \ref{fig:steps}.
As the fine-tuning progresses, there is a notable improvement in both MC1 and MC2 accuracy on TruthfulQA, with the metrics stabilizing around the 1000th step.
Conversely, the performance on ARC-C, HellaSwag, and MMLU almost maintains consistency throughout fine-tuning.
Therefore, we adopt 1000 fine-tuning steps in our experiments.

\subsection{Towards Understanding How to Learn Truthfulness
in Gradual Self-Truthifying}
\label{sec:appendix_understand_gradual}
— \textit{How does the number of iterations affect the model truthfulness as well as other fundamental abilities?}

\begin{figure}[ht]
    \centering
    \includegraphics[width=0.8\linewidth]{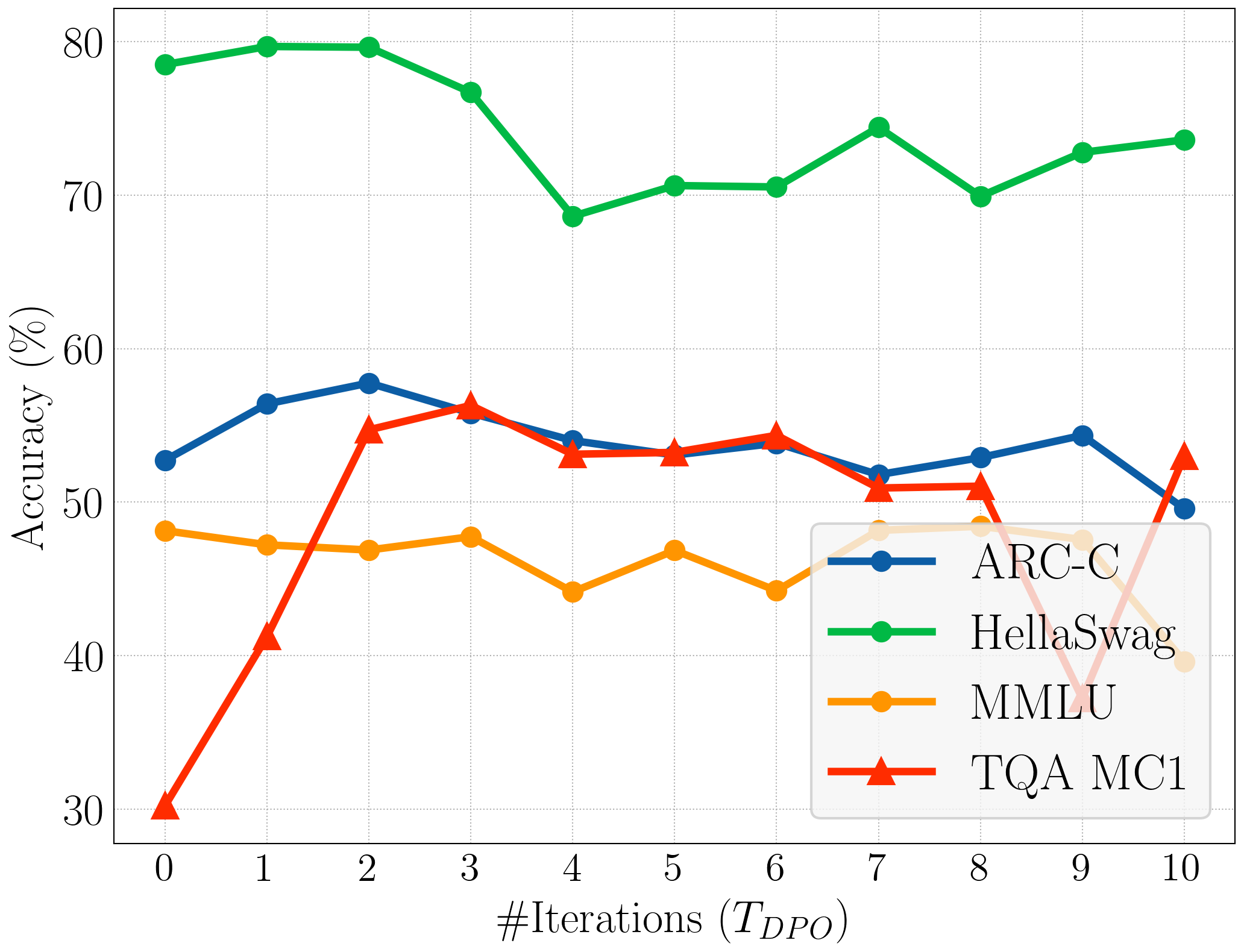}
    \caption{Performance of \name on four benchmark datasets with varying iterations. Note that the performance on TruthfulQA's MC2 task is not reported since MC2 accuracy becomes NaN at the third iteration.}
    \label{fig:iterations}
\end{figure}

Figure \ref{fig:iterations} displays the changes in model performance on benchmark tasks across iterations. Initially, within the first few iterations, there is a significant improvement in the model performance on TruthfulQA's MC1 task, while the results on other datasets remain relatively stable. However, beyond a certain number of iterations, a decline in performance is observed across all datasets. Notably, the MC2 accuracy on TruthfulQA registers as NaN starting from the third iteration, which is not presented in the figure.
These findings suggest that, on one hand, \name effectively enhances performance on TruthfulQA within just a few iterations without detrimentally affecting other capabilities. On the other hand, extending the number of iterations leads to a deterioration in overall performance, likely due to the overfitting issue associated with DPO, as discussed in Section \ref{sec:gradual}. 
Consequently, we set $T_{DPO} = 2$ (equivalent to $T=1$) for our experimental setup.

— \textit{How does the choice of reference model in DPO affect the performance of the learned model?}
In \name, we set the reference model as the current base model when applying DPO. But, an alternative exists—using the pretrained model as the reference model, which means that the reference model in DPO is fixed across iterations. 

\begin{figure}[ht]
    \centering
    \includegraphics[width=0.8\linewidth]{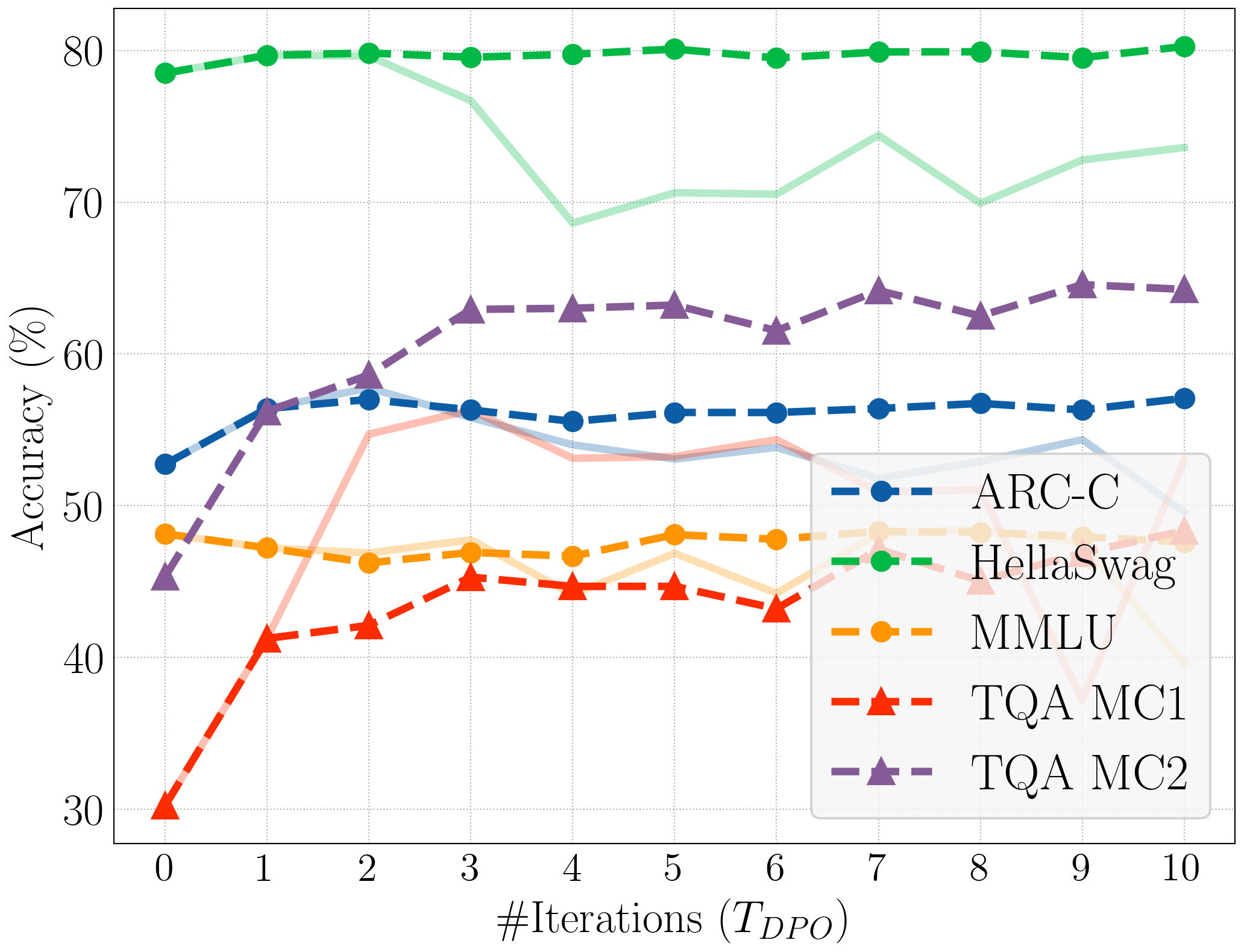}
    \caption{Performance of \name on four benchmark datasets with varying iterations. The dotted lines indicate using the pretrained model as the reference model in DPO while the transparent lines indicate using the base model as the reference model in DPO.}
    \label{fig:ref_model}
\end{figure}

To explore this, we fix the reference model as the pretrained model, execute \name for 10 iterations, and assess the performance on benchmark datasets, as depicted in Figure \ref{fig:ref_model}. The transparent lines represent the scenario where the current base model serves as the reference (corresponding to Figure \ref{fig:iterations}’s results), whereas the dotted lines symbolize the use of the pretrained model as the reference.
We observe that the results on ARC-C, HellaSwag, and MMLU remain steady, suggesting a preservation of the model's core capabilities. 
On the other hand, there is a consistent enhancement in model performance on TruthfulQA, indicative of a gradual improvement in model truthfulness. Note that MC2 accuracy is omitted in Figure \ref{fig:iterations} due to the potential non-applicability (NaN); however, with the pretrained model as the reference, MC2 accuracy won't become NaN but instead exhibits an increase.
Despite this stable growth, it's noticed that the optimal performance on TruthfulQA does not reach the levels achieved when the current base model is used as the reference.

The results are reasonable since if using the pretrained model as the reference model, DPO's objective function will apply a regulation factor (\textit{i.e.}, $\beta$) to limit deviations from the pretrained model. Consequently, compared to the scenario where the current base model is the reference, the learned model is more similar to the pretrained model, yielding stable results across ARC-C, HellaSwag, and MMLU. Nevertheless, this regulatory effect also means that gains in truthfulness are slower and less pronounced.

In summary, if a model trainer cares more about the influence on the model's core capabilities and there is less concern for rapid and significant enhancements in truthfulness, then the pretrained model should serve as the reference. Conversely, if a model trainer prioritizes efficiency and substantial gains in truthfulness—and is willing to accept minor impacts on core capabilities—then opting for the current base model as the reference and executing \name for only a few iterations emerges as a highly efficient and effective approach.

\section{Future Directions}
\label{sec:future}
Despite the advancements within the proposed method, there are still several questions remaining open for exploration. Here, we offer some corresponding potential solutions.
1) As demonstrated in Section \ref{sec:gradual} and \cite{azar2023general}, DPO might over-fit to the pairwise preference data. A simple solution is early-stopping. Besides, \cite{azar2023general} proposes a strategy named IPO to solve it.
2) While our focus here is truthfulness, LLMs are expected to exhibit a range of desirable qualities.
A natural question arises: \textit{How can we enhance LLMs' multiple capabilities simultaneously?} A naive idea would be extending the creation procedure in \name to generate pairwise answers that differ in diverse attributes (\textit{e.g.}, harmful \textit{v.s.} harmless, moral \textit{v.s.} immoral).
Overall, numerous intriguing questions are awaiting our exploration in the future.

\end{document}